\documentclass[lettersize,journal]{IEEEtran}
\usepackage{makecell}
\usepackage{amsmath}
\usepackage{enumitem}
\usepackage{amsfonts}
\usepackage{latexsym}
\usepackage{graphicx}
\usepackage{listings}
\usepackage{algpseudocode}
\usepackage{subfig}
\usepackage{caption}
\usepackage{bm}
\usepackage{booktabs}
\usepackage{color}
\usepackage{mathtools}
\usepackage{todonotes}
\usepackage{multirow}
\usepackage{dsfont}
\usepackage{amsthm}

\newcommand{\etal}{\textit{et al. }}
\usepackage[linesnumbered,ruled,vlined]{algorithm2e}

\begin{document}

\title{HiBid: A Cross-Channel Constrained Bidding System with Budget Allocation by Hierarchical Offline Deep Reinforcement Learning}

\author{Hao Wang*, Bo Tang*, Chi Harold Liu,~\IEEEmembership{Senior member,~IEEE}, Shangqin Mao, Jiahong Zhou, Zipeng Dai, Yaqi~Sun, Qianlong Xie, Xingxing Wang, Dong Wang

\thanks{H. Wang and B. Tang contributed equally to this work. H. Wang, C. H. Liu (Corresponding Author), and Z. Dai are with School of Computer Science and Technology, Beijing Institute of Technology, Beijing 100081, China. Email: liuchi02@gmail.com}
\thanks{B. Tang is with Meituan and Institute for Advanced Algorithms Research, Shanghai, China.}
\thanks{S. Mao, J. Zhou, Y. Sun, Q. Xie, X. Wang and D. Wang are with Meituan, Beijing, China.}
}


\maketitle

\begin{abstract}
Online display advertising platforms service numerous advertisers by providing real-time bidding (RTB) for the scale of billions of ad requests every day. The bidding strategy handles ad requests cross multiple channels to maximize the number of clicks under the set financial constraints, i.e., total budget and cost-per-click (CPC), etc. Different from existing works mainly focusing on single channel bidding, we explicitly consider cross-channel constrained bidding with budget allocation. Specifically, we propose a hierarchical offline deep reinforcement learning (DRL) framework called ``HiBid'', consisted of a high-level planner equipped with auxiliary loss for non-competitive budget allocation, and a data augmentation enhanced low-level executor for adaptive bidding strategy in response to allocated budgets. Additionally, a CPC-guided action selection mechanism is introduced to satisfy the cross-channel CPC constraint. Through extensive experiments on both the large-scale log data and online A/B testing, we confirm that HiBid outperforms six baselines in terms of the number of clicks, CPC satisfactory ratio, and return-on-investment (ROI). We also deploy HiBid on Meituan advertising platform to already service tens of thousands of advertisers every day.
\end{abstract}


\section{Introduction}
Real-time systems are witnessing an ever-growing significance in our society, including Industrial Internet of Things (IIoT) systems employed for crowdsensing \cite{rt_1}, online cloud auction systems that facilitate seamless streaming services \cite{rt_2,rt_3}, and real-time bidding (RTB) systems that have emerged as an indispensable component in modern online E-commerce, by offering real-time bidding services~\cite{intro_RTB1} for millions of advertisers participating in online display advertising. 
RTB creates an advertisement (ad) auction that varies depending on the platform, to allow interested advertisers to bid simultaneously for ad impressions. Different auction mechanisms (e.g., generalized first price in Google AdSense and Vickrey-Clarke-Groves in Facebook) and pricing mechanisms (e.g., cost-per-mille and cost-per-time) are chosen based on advertising and service format on the platform. One common type is the Generalized Second Price (GSP \cite{GSP}) auction with cost-per-click (CPC \cite{CPC}) pricing, which charges advertisers the second-highest bidding price if a user clicks on their ad. It provides a fair and flexible advertising format with an efficient evaluation of advertising performance for advertisers, which plays an important role in the online advertising industry.

\begin{figure}[tb]
\centering
\includegraphics[width=0.98\columnwidth]{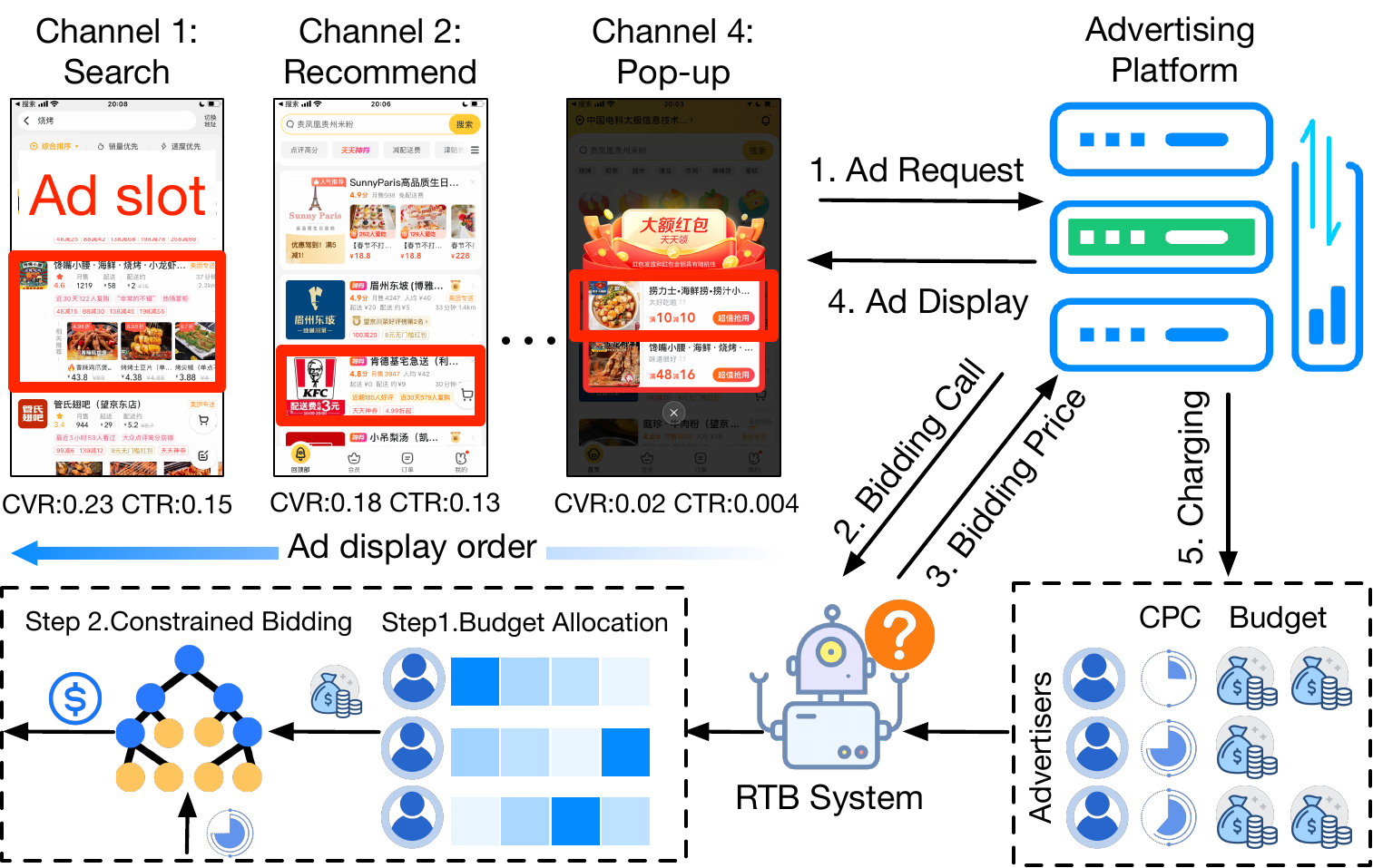}
\caption{Overview of the considered cross-channel constrained bidding (c$^3$-bidding) scenario.}
\label{fig:cover}
\end{figure}

As shown in Fig.~\ref{fig:cover}, for better use of limited budget within a financial constraint like CPC, more advertising platforms have begun to offer automated bidding services across various ad channels, such as recommendation ads (i.e., ads that recommend products to a potential user), search ads (i.e., ads shown in response to user search content), etc. 
Different ad channels have varying numbers of ad impressions and users on the basis of their behavioral habits. Some high-quality channels, which are reflected by higher conversion rate (CVR) and click-through rate (CTR), could bring better advertising effectiveness and thus revenue as a return.

It is essential to allocate budget to different channels because: (a) from advertisers' perspective, due to the inconsistent peak time and volume of ad requests in different channels, investing more budget in those channels (which are more suitable for themselves) can potentially avoid excessive consumption in other channels, leading to higher ROI; and (b) from the platform's perspective, due to the limited available ad space, appropriately allocating advertisers' budgets can help advertisers more precisely pinpoint their target users across different channels, thereby enhancing ad relevance and effectiveness. Advertisements with higher efficacy draw advertisers to increase their investments in ad campaigns, leading to a growth in the platform's revenue, which is a mutually beneficial outcome; and (c) from a market perspective, owing to the higher cost-effectiveness of high-quality channels for all advertisers, distributing the total budget on each channel in a reasonable way may prevent advertisers from competing for those high-quality channels simultaneously.

Extensive works have been conducted on constrained bidding. Most of them focused on improving bidding strategy in a single channel under the set budget \cite{constrained_bidding2,cbrl} but did not adjust their bidding or budget allocation strategy across all channels. Hence, they cannot scale well to cross-channel bidding problems. 
Previous works on budget allocation \cite{budget1,budget2} derived the optimal strategy using a prediction model to forecast the expected return of the bidding strategy. However, the bidding and allocation strategies are ``stand-alone", that there exist many dynamic factors to affect the performance of the RTB system, causing fluctuations of budget allocation and bidding price. 
Therefore, an insight is that integrating these two may receive mutual feedback to bring better advertising results to advertisers.

In this paper, we explicitly consider the problem of \textbf{c}ross-\textbf{c}hannel \textbf{c}onstrained \textbf{b}idding with budget allocation, ``c$^3$-bidding'', where the key challenges are: (a) advertisers may allocate most of their budgets to high-quality channels, leading to possible contentions and a decrease in overall performance; (b) the advertisers' daily budget and the platform allocated budget on each channel highly likely change over time, and therefore the bidding strategy should dynamically adapt to the changing budget constraint; and (c) since each channel is bid independently, it is challenging to ensure the budget constraint for individual channels while  guaranteeing cross-channel CPC constraints.

In practice, billions of ad requests arrive sequentially for tens of thousands of advertisers, and thus the solution space of the considered c$^3$-bidding problem is huge which is not solvable by using optimization methods \cite{or}. Meanwhile, the rapid advancement of offline deep reinforcement learning (DRL) demonstrates its potential for learning an optimal policy as well as satisfying the given constraint from the large-scale data.
Furthermore, we observe that there is a hierarchy of budget allocation and constrained bidding in RTB. That is, the former assigns a percentage of the budget to each channel from market perspective, while the latter cares more about the suitable bidding price to win the ad impression opportunities under the allocated budget. Thus, we model the considered c$^3$-bidding problem as a hierarchical Constrained Markov Decision Process (CMDP \cite{CMDP}). Existing hierarchical approaches \cite{opal,HRL_ADD} neither address the cross-channel CPC constraints nor consider the decline in performance due to inappropriate allocation.
To this end, we propose a novel hierarchical offline DRL framework called ``HiBid'' based on the state-of-the-art offline DRL approach MCQ \cite{mcq}, as the start point of design. Our contribution is three-fold:
\begin{enumerate}
    \item We propose ``HiBid'', a hierarchical DRL framework for c$^3$-bidding problem which maintains a high-level planner for budget allocation and a low-level executor for cross-channel constrained bidding. 
    \item We introduce batch loss \cite{crossdqn} for budget allocation to prevent over-allocation on specific channels, $\lambda$-generalization \cite{bcorle} for constrained bidding to adaptively respond to changing budget. Then, we propose a CPC-guided action selection mechanism to significantly improve the cross-channel CPC satisfactory ratio, which also has wider applicability to other metrics as well.
    \item We conduct extensive experiments on a large-scale real dataset in Meituan advertising platform. Results show that HiBid outperforms six baselines in terms of number of clicks, CPC satisfactory ratio and ROI. We also deployed HiBid, and performed online A/B testing to validate its effectiveness.
\end{enumerate}

The rest of the paper is organized as follows. We review the related work in Section \ref{sec-related} and present the system model in Section \ref{sec-system}. We introduce preliminaries in Section \ref{sec-pre}. We propose HiBid in Section \ref{sec-solution}, followed by the experimental results in Section \ref{sec-exp}. Finally, we conclude the paper in Section \ref{sec-con}.

\section{Related Work}
\label{sec-related}
\subsection{Real-Time Bidding (RTB) systems}
RTB attracts much attention and has been widely studied for various applications \cite{RTB2}. Some efforts have been devoted to designing bidding mechanisms to enhance the effectiveness and fairness of advertising auctions from the platform perspective. For example, Zhou \etal in \cite{price1} introduced a novel deep distribution network for optimal bidding in both open and closed online first-price auctions. Zhang \etal in \cite{price2} proposed a succinct and effective bid shading algorithm without parametric assumptions for the win distribution. Ren \etal in \cite{price3} proposed a comprehensive framework to jointly optimize user response prediction and bid landscape forecasting. Furthermore, there have been studies that approach the optimization of bidding strategies from the advertiser perspective, to improve the effectiveness of their ads during auctions. For example, Wu \etal in \cite{drlb} developed a model-free DRL framework ``DRLB'' for constrained bidding to cope with the volatility of the auction environment. Yang \etal in \cite{pid} abstracted the essential demand of advertisers in RTB and proposed an effective linear programming solution.
Those works focused on optimizing bidding prices under the given constraint for a single ad channel, to adapt to the unpredictability of the auction environment and satisfy advertisers' requirements. However, there exist multiple advertising channels with significant quality differences in practical deployment. Also, some studies developed ways to allocate budget across multiple ad channels given a total budget constraint, e.g., some works \cite{budget1,budget2} formalized it as well-stated optimization problems. These methods required accurate estimation of outcome distributions (e.g., the expected number of clicks from choosing a particular budget), which is impracticable in a dynamic auction environment. 

Different from the above works, this paper explicitly focuses on simultaneous budget allocation and constrained bidding for multiple ad channels, to maximize the advertising effectiveness for advertisers while ensuring platform revenue.

\subsection{Deep Reinforcement Learning (DRL)} 
DRL has been widely applied in real-time systems, including user-item recommendation \cite{rec_user_item1}, ad-slots allocation \cite{rec_ad1}, and real-time bidding for ad impression auctions \cite{constrained_bidding1,constrained_bidding2,cbrl,uscb}. 
Cai \etal in \cite{constrained_bidding1} and Zhao \etal in \cite{bid_zhao} utilized DRL to learn the optimal bid for a single ad in display advertising and sponsored search, respectively. He \etal in \cite{uscb} formulated the budget and financial constraints simultaneously and leveraged DRL to find a unified optimal bidding function on behalf of an advertiser. Unfortunately, the optimal bidding function may not yield optimal results in uncertain auction markets. Wang \etal in \cite{cbrl} proposed a curriculum-guided Bayesian DRL method to generalize to highly dynamic ad markets with ROI constraints. 
However, the mentioned DRL-based works only focus on single ad channel bidding and have not considered joint modeling across multiple channels. Due to the joint constraint settings of the advertiser's financial requirements across channels, bidding individually cannot yield optimal results. It is insightful to model the relationship between channels jointly for bidding through a unified approach. Therefore, we leverage hierarchical reinforcement learning to jointly model cross-channel bidding and adjust bidding strategies through high-level budget allocation, which is one motivation of our work.

DRL provides a promising approach to address the c$^3$-bidding problem by interacting with the environment and updating policy iteratively. However, it is not suitable for training the agent in an online setting due to the potential financial risk involved. In offline DRL, the agent learns from a fixed dataset of past interactions, rather than learning online in real-time. The main challenge of offline DRL is the distribution shift \cite{distributionshift} of state-action visitation between the learned policy and behavior policy. Recent work \cite{offline_penalities1} utilized distributional penalties to regularize the learned policy to stay close to the behavior policy. Other methods \cite{bcq} used generative models to approximate the behavior distribution to stay within the support of offline data during the value back up. Ajay \etal in \cite{opal} proposed a hierarchical offline reinforcement learning method with unsupervised primitive extraction. 
However, directly applying existing offline DRL algorithms may not effectively solve our considered c$^3$-bidding problem, as there is no effective method to solve the cross-channel CPC constraint and the changing allocated budget.

Constrained DRL focuses on designing efficient algorithms to find optimal policies for CMDP problems under the given constraints \cite{constraint1}. Some works converted the CMDP problem into a Lagrangian dual problem \cite{lagrangian}, and then found an optimal Lagrangian multiplier $\lambda$ as well as the corresponding policy which satisfied the constraint. Here $\lambda$ can be manually adjusted as a hyperparameter, which is policy-sensitive and hard to fine-tune. In recent works, gradient descent \cite{lambda_sgd} or bisection search \cite{lambda2} were developed to get the optimal value of $\lambda$. Unfortunately, the policy needs to be retrained every time when the value of $\lambda$ changes until the constraint is satisfied. The iterative training process is unacceptable in c$^3$-bidding problem due to the frequently changed budget that brings huge computational overhead. Thus, in this paper we adopt a $\lambda$-generalization \cite{bcorle} method to learn diversified bidding strategies that can dynamically respond to the changed budget constraint. Nevertheless, for the cross-channel CPC constraint, we need a more appropriate way to solve it, which is the key contribution of this paper as CPC-guided action selection in Section \ref{sec-action-selection}.

\begin{table}[tb]
	\centering
	\caption{Important notations used in this paper.}\label{tab:notation}
	\begin{tabular}{l p{5.5 cm}}
		\toprule
		Notation & Explanation \\
		\toprule
		\ $P, p$ & The total number, index of channels\\
		\hline\rule{0pt}{8pt}
  	    $M, m$ & The total number, index of advertisers\\
		\hline\rule{0pt}{8pt}
		$I_p, i$ & The total number, index of ad requests on channel $p$\\
            \hline\rule{0pt}{3pt}
            $B_m,CPC_m^{set}$ & Total budget and CPC constraint set by advertisers $m$ \\
            \hline\rule{0pt}{8pt}
            $Click(\cdot), Cost(\cdot)$ & The number of clicks and actual cost\\
            \hline\rule{0pt}{8pt}
            $a_{m,p,i}$ & Bidding price offered by advertiser $m$ for request $i$ on channel $p$ \\
            \hline\rule{0pt}{8pt}
            $s_p^h,a_p^h,r_p^h,c_p^h$ & State, action, reward and cost for channel $p$ in high-level MDP \\
            \hline\rule{0pt}{8pt}
            $s_{i}^l,a_{i}^l,r_{i}^l,c_{i}^l$ & State, action, reward and cost of ad request $i$ in low-level MDP \\
            \hline\rule{0pt}{8pt}
            $Q^{(\cdot)}_{\theta},Q^{(\cdot)}_{\theta'},\hat{Q}^u_{\eta}$, $\hat{Q}^c_{\phi} $ &  Q-network, target network and evaluation networks\\
            \hline\rule{0pt}{8pt}
            $\lambda,\lambda^*_i$& Lagrangian multiplier in bidding strategy, optimal $\lambda$ for ad request $i$\\
            \hline\rule{0pt}{8pt}
            $N, N_p$& Data repetition times in offline training and sample number of $\lambda$ in online prediction\\
            \hline\rule{0pt}{8pt}
             $\mathcal{L}_{(\cdot)}, w_1,w_2,w_{b}$& Loss functions, weight of Q-function loss, OOD action loss, and batch loss \\
             \hline\rule{0pt}{8pt}
            $Impr, Click$ & The total number of impressions and clicks  \\
            \hline\rule{0pt}{8pt}
            $ROI, CPC, CSR$ & Return-on-investment, cost-per-click, and CPC satisfactory ratio \\
		\bottomrule
\end{tabular}
\vspace{-5mm}
\end{table}
\vspace{-5mm}

\section{System Model}
\label{sec-system}
In this paper, our overall objective is to maximize the total ad clicks while satisfying all the advertisers' set budget and CPC constraints, while ensuring that the platform's revenue remains within an acceptable range. Without loss of generality, we consider an advertising platform is servicing $M$ advertisers across $P$ ad channels with $I_p$ ad requests on channel $p$ in a day. Each advertiser $m$ sets a daily budget $B_m$ (i.e., the maximum amount of money they are willing to spend for their advertising campaign) and expected cost-per-click $CPC_m^{set}$, then the overall objective is formulated as:
\begin{align}
 \underset{{a_{m,p,i}}}{\mathrm{maximize}} &\sum\limits_{m=1}^{M}\sum\limits_{p=1}^P\sum\limits_{i=1}^{I_p} Click(a_{m,p,i})\label{overall-objective} \\
  \mathrm{subject\,to:\,} &\left|\sum_{m=1}^M  Cost(a_{m,p,i}) - \kappa_p \right| \leq \epsilon,  \forall p \in \{1,\dots,P\}, \label{complete-problem-batch-constraint} \\
   & \frac{\sum_{p=1}^{P}\sum_{i=1}^{I_p} Cost(a_{m,p,i})}{\sum_{p=1}^{P}\sum_{i=1}^{I_p} Click(a_{m,p,i})} \leq CPC_m^{set}, \\
   & \sum_{p=1}^P\sum_{i=1}^{I_p} Cost(a_{m,p,i}) \leq {B_m},  \forall m \in \{1,\dots,M\},  
\end{align}
where $a_{m,p,i}$ represents the bidding price offered by an advertiser $m$ for a request $i$ on a channel $p$. This bidding price indicates the amount an advertiser $m$ is willing to pay to display their ad in response to the request $i$ on that particular channel. ${Cost}(a_{m,p,i})$ corresponds to the actual expense spent by advertiser $m$ when their specific bid for request $i$ is successful. $Click(a_{m,p,i})$ indicates whether or not a user clicks on the ad after it is displayed. If the offered bidding price does not win the auction, both ${Cost}(a_{m,p,i})$ and ${Click}(a_{m,p,i})$ are set to $0$. The constraint in Eqn. (\ref{complete-problem-batch-constraint}) is added to prevent the channels' revenue from too much fluctuation. $\epsilon>0$ is a constant, representing the acceptable fluctuation range of the platform. $\kappa_p$ is calculated by multiplying historical CTR, CPC and impression count, that represents the expected consuming capacity in an ideal situation.
    
\begin{figure}[tb]
\centering
\includegraphics[width=0.8\columnwidth]{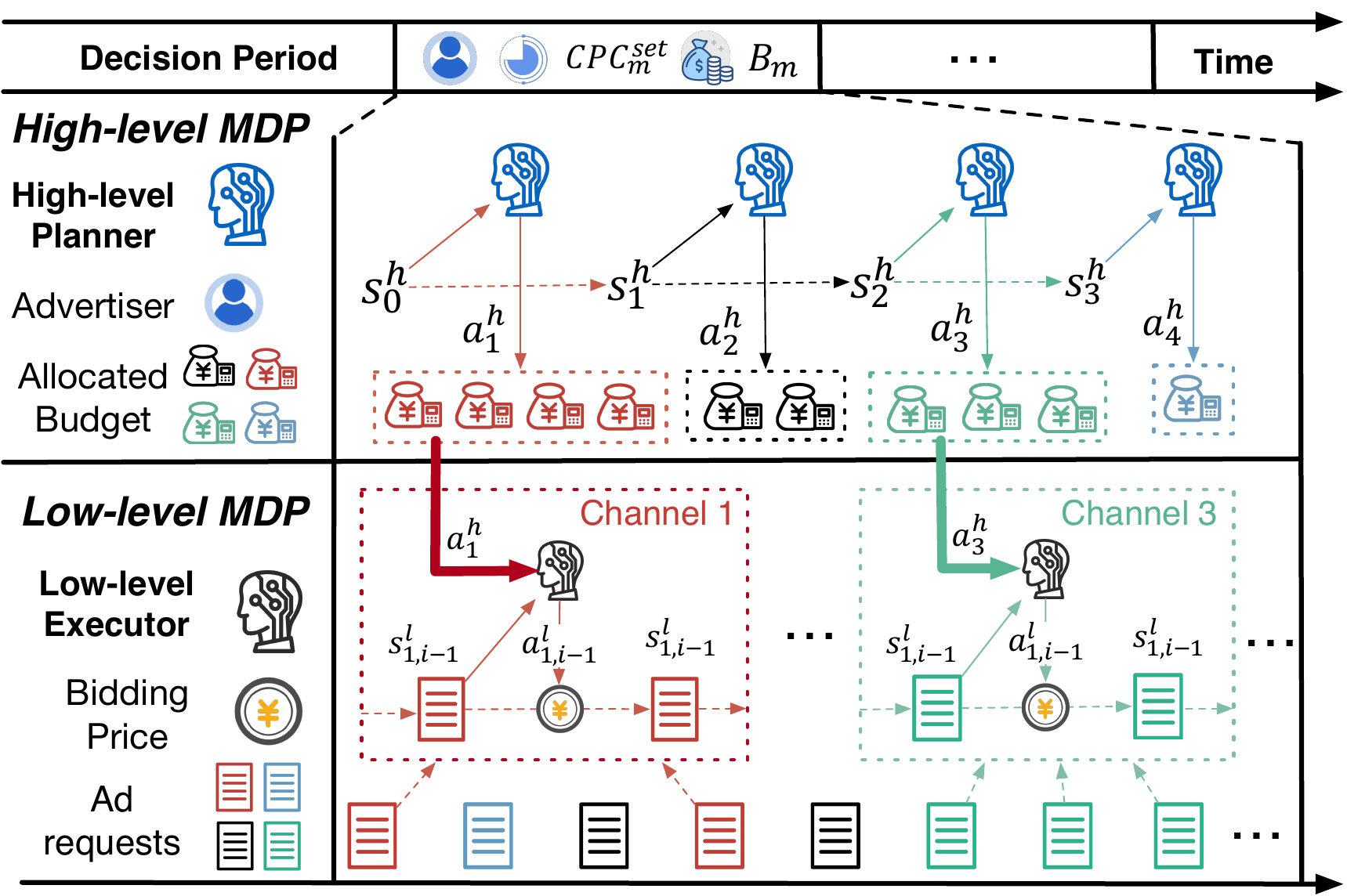}
\caption{Hierarchical CMDP modeling for c$^3$-bidding.}
\label{fig_mdp}
\end{figure}

In practice, the incoming ad requests of all advertisements are not known as a priori. This makes it hard to employ traditional combinatorial optimization methods to solve the considered $c^3$-bidding problem. Considering its inherent hierarchical nature, we can allocate budgets to all the advertisers from the platform's perspective. For each advertiser, bids can be made based on the allocated budget and set financial constraints. The budget allocation for advertisers and the decision-making of bidding prices for sequentially incoming ad requests both exhibit Markovian properties.
Therefore, we model the c$^3$-bidding problem as a hierarchical CMDP where high-level and low-level MDPs are executed on different timescales. As shown in Figure \ref{fig_mdp}, the high-level MDP is responsible for allocating the budget at intervals, while the low-level MDP bids for each ad request according to the allocated budget.

\subsection{High-level MDP for Budget Allocation}
The high-level planner needs to allocate the budget to maximize the number of user clicks while ensuring the revenue on each channel stays within an acceptable range. Thus the objective of the high-level planner is:
\begin{align}
 \underset{{a_{m,p}^h}}{\mathrm{maximize}} &\sum\limits_{m=1}^{M}\sum\limits_{p=1}^P Click^h(a^h_{m,p})\label{high-objective} \\
  \mathrm{subject\,to:\,} & \sum_{p=1}^P a^h_{m,p} \leq {B_m}, \quad \forall m \in \{1,\dots,M\}, \label{high-budget}  \\
 \quad&\left|\sum_{m=1}^M  Cost^h(a^h_{m,p}) - \kappa_p \right| \leq \epsilon, \quad \forall p \in \{1,\dots,P\}, \label{batch-constraint}
\end{align}
where $a_{m,p}^h$ denotes the allocated budget on channel $p$ for advertiser $m$, $Click^h(a_{m,p}^h)$ and $Cost^h(a_{m,p}^h)$ represent the number of clicks and cost given the budget $a_{m,p}^h$ within the interval, respectively. Meanwhile, the revenue constraint in Eqn. (\ref{batch-constraint}) is also regarded as a channel-capacity constraint that prevents advertisers from engaging in severe competition for high-quality channels. 
%


We formulate the high-level MDP for each advertiser $m$ as a tuple $ (\mathcal{S}^h,\mathcal{A}^h,\mathcal{T}^h,\gamma^h,\mathcal{R}^h,\mathcal{C}^h)$. That is, the high-level planner allocates the channel-level budget at each interval according to the advertiser's budget requirements, while each channel allocation is regarded as a decision step. The details are defined as follows:

\textbf{State $\mathcal{S}^h$}: Let $\mathcal{S}^h$ denote the higher-level state space, consisting of the allocated budget and the historical statistics on the advertiser-level (e.g., the average CTR and CVR).

\textbf{Action $\mathcal{A}^h$}: The action $a_{p}^h\in \mathcal{A}^h$ is the budget assigned to channel $p$. We discretize the action by using the percentage of the budget and mask invalid actions that go beyond the total budget.

\textbf{Reward $\mathcal{R}^h$}: Reward $r^h_p \in \mathcal{R}^h$ is defined as the sum of number of user clicks in channel $p$ within the interval. 

\textbf{Constraint $\mathcal{C}^h$}: The channel-capacity constraint and total budget constraint are given in Eqn. (\ref{high-budget}) and Eqn. (\ref{batch-constraint}), respectively.

\subsection{Low-level MDP for Cross-Channel Constrained Bidding}
After receiving the allocated budget $a^h_p$ on each channel, the low-level executor aims to maximize the number of clicks while satisfying that budget and CPC constraint simultaneously. For each advertiser $m$, the objective function of the low-level executor can be formulated as:
\begin{align}
    \underset{a^l_{p,i}}{\mathrm{maximize}} &\sum\limits_{p=1}^{P}\sum\limits_{i=1}^{I_p} Click(a^l_{p,i}) \label{low-objective}\\
     \mathrm{subject\,to:\,} &\sum_{i=1}^{I_p} Cost(a^l_{p,i}) \leq a^h_{p},\quad \forall p \in \{1,\dots,P\},  \label{low-budget-constraint}\\
     & \frac{\sum_{p=1}^{P}\sum_{i=1}^{I_p} Cost(a^l_{p,i})}{\sum_{p=1}^{P}\sum_{i=1}^{I_p} Click(a^l_{p,i})} \leq CPC_m^{set} \label{CPC-constraint},
\end{align}
where $Cost(a^l_{p,i})$ and $Click(a^l_{p,i})$ denotes the real cost and whether the user clicks the ad after giving a bidding price $a^l_{p, i}$, respectively.
Each request $i$ comes from a specific channel and only affects the cost of that channel, and thus we model each channel individually. Then, the CMDP for channel $p$ can be formulated as a tuple $(\mathcal{S}^l,\mathcal{A}^l,\mathcal{T}^l,\gamma^l,\mathcal{R}^l,\mathcal{C}^l)$, which is defined as follows:

\textbf{State $\mathcal{S}^l$}: The low-level state space is a collection of allocated budgets $a^h_p$, request and advertiser level information. The request-level information includes time, and current advertising status (e.g., budget consumption rate and financial constraints satisfactory ratio). The advertiser-level information is identical to the high-level planner state.

\textbf{Action $\mathcal{A}^l$}: Following \cite{uscb}, an action $a_i^l \in [a^l_{\min},a^l_{\max}]$ represents the bidding ratio and the final bidding price is calculated by $a_i^l*CPC_m^{set}$. 

\textbf{Reward $\mathcal{R}^l$}: For each request $i$, reward $r_i^l\in \{0,1\}$ is set to $1$ if the bidding is successful and the user eventually clicks the ad.

\textbf{Constraint $\mathcal{C}^l$}: Single-channel budget constraints and cross-channel CPC constraints are given in Eqn. (\ref{low-budget-constraint}) and Eqn. (\ref{CPC-constraint}), respectively.



\section{Preliminary}
\label{sec-pre}
In order to apply offline DRL methods to the considered c$^3$-bidding problem, the key is to maintain a conservative value estimation (i.e., to eliminate the possible over-estimation). Recall that Q-network $Q_\theta(s,a)$ measures the accumulative discounted reward starting from state-action pair $(s,a)$ parameterized by $\theta$. $Q_\theta(s,a)$ can be improved via minimizing the temporal difference \cite{dqn} as:
\begin{align}
\label{loss-q}
      \mathcal{L}_Q(\theta)= \mathbb{E}_{(s,a,r,s')\sim\mathcal{D}}\left[(r+\gamma Q_{\theta'}(s',a')-Q_\theta(s,a))^2\right],
\end{align}
where $a'=\mathop{\arg \max}_{a'}Q_{\theta'}(s',a')$ and $Q_{\theta'}$ is a target network for learning stability. The out-of-distribution (OOD) state-action pairs bring extrapolation error \cite{bcq} during the offline training, resulting in a severely overestimated value function and an aggressive bidding strategy. This strategy may result in a catastrophic financial loss in c$^3$-bidding because it is inclined to give higher bidding prices, leading to vicious competition and significant risks. Thus it is important to keep conservatism in value estimation which can help prevent the learned policy from taking risky actions.

In this paper, we adopt Mildly Conservative Q-learning (MCQ \cite{mcq}) as the start of the design for both the high-level planner and the low-level executor, where OOD state-action pairs are actively trained by assigning proper pseudo Q values. In the considered c$^3$-bidding problem, the policy distribution within the dataset exhibits a multi-modality pattern due to the highly non-stationary external market. A simple parameterized approach (e.g., using MLP with cross-entropy) cannot work well as it focuses on mapping input to output and neglect the full distribution. Specifically, we utilize a conditional variational autoencoder (CVAE \cite{cvae}) to extensively model the distribution of the behavior policy $\mu$. Given log data, the objective of CVAE is to reconstruct actions conditioned on the states, such that the generated actions come from the same distribution as the actions in the log. The utilized CVAE is denoted as $G_\omega(s)$ parameterized by $\omega$, which is consisted of an encoder $G^E_{\omega_1}(s,a)$ and an decoder $G^D_{\omega_2}(s,z)$. We optimize its variational lower bound by:
\begin{equation}
\small
\label{loss-vae}
    \mathcal{L}_{CVAE}(\omega) = \mathbb{E}\left[\big(a-G^D_{\omega_2}(s,z)\big)^2+KL\big(G^E_{\omega_1}(s,a),\mathcal{N}(0,\mathbf{I})\big)\right],
\end{equation}
where hidden state $z= G^E_{\omega_1}(s,a)$, $KL(\cdot)$ denotes the KL-divergence, $\mathcal{N}$ is multivariate normal distribution, and $\mathbf{I}$ is the identity matrix.

Then, given a state $s$, we generate several in-distribution actions $a_\mu$ by CVAE $G_\omega(s)$, and the auxiliary loss for OOD actions is calculated by:
\begin{align}
\label{loss-ood}
    \mathcal{L}_{OOD}(\theta)= \mathbb{E}_{s\sim\mathcal{D}}[(\max_{a_{\mu}\sim G_\omega(s)} Q_{\theta}(s,a_{\mu}) - Q_{\theta}(s,a_{\pi}))^2].
\end{align}
If $a_\pi$ (generated by current policy $\pi$) is an OOD action, the auxiliary loss will limit the corresponding value estimation below the maximum value of in-distribution action $a_\mu$. In this way, we help the value estimator stay conservative such that OOD actions will not be severely overestimated. However, MCQ cannot guarantee the budget allocation and bidding strategy to satisfy user requirements, hence we explicitly propose three key modules. 


\begin{figure}[tb]
\centering
\includegraphics[width=\columnwidth]{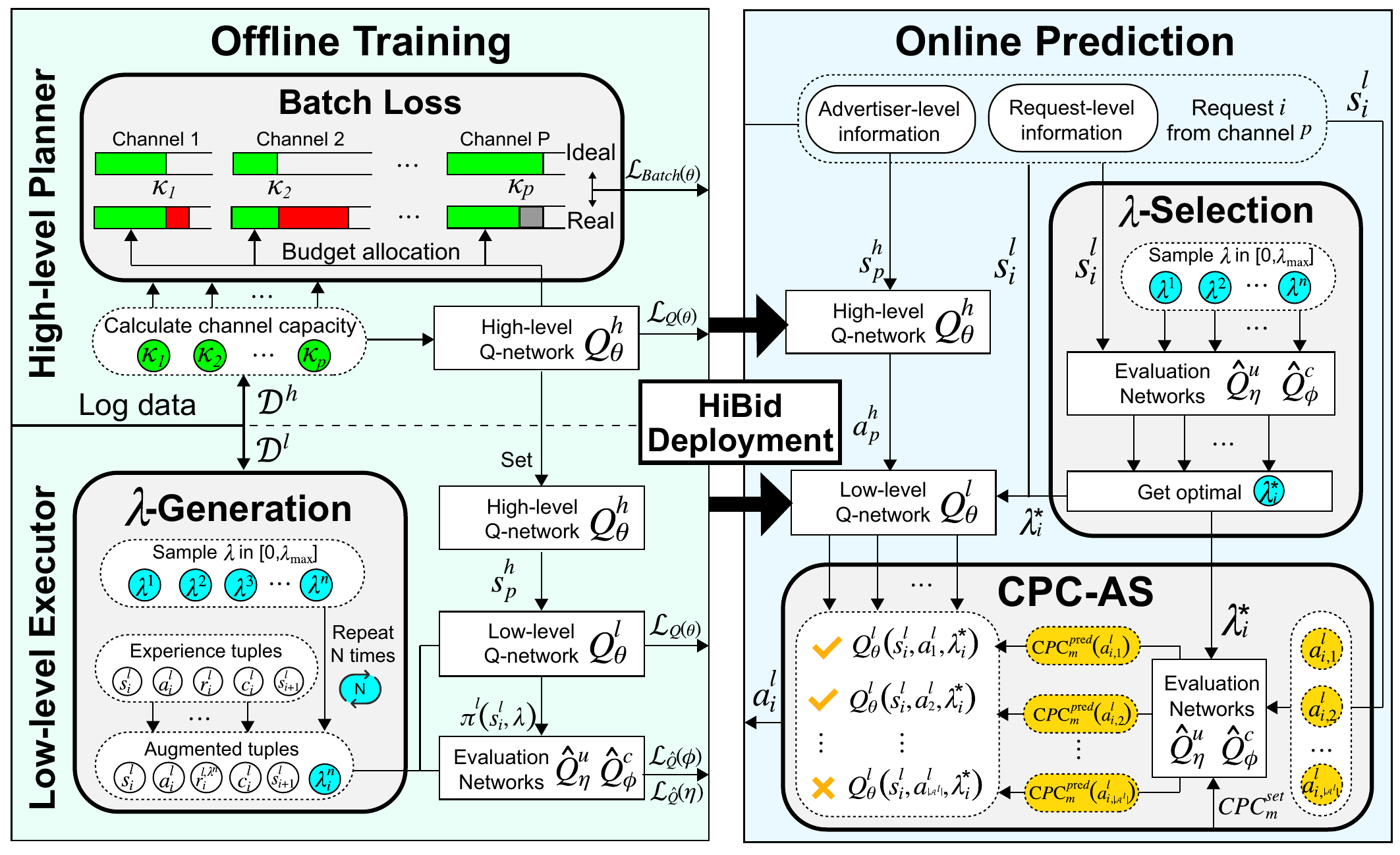}
\caption{The proposed framework: HiBid}
\label{fig-solution}
\end{figure}

\section{Our solution: HiBid}
\label{sec-solution}
Due to the large solution space and multiple constraints of the considered c$^3$-bidding problem, we propose a hierarchical offline DRL framework called HiBid, as shown in Figure \ref{fig-solution}. We first introduce an auxiliary batch loss \cite{crossdqn} to prevent over-allocating budget on specific channels (see Section \ref{sec-batch-loss}), $\lambda-$generalization \cite{bcorle} for constrained bidding to adaptively respond to changing budget (see Section \ref{sec-lambda-generation}), and a CPC-guided action selection (CPC-AS) scheme for the cross-channel CPC constraint satisfaction (see Section \ref{sec-action-selection}).

\subsection{Auxiliary Batch Loss for Non-competitive Budget Allocation}
\label{sec-batch-loss}
Recall that the high-level planner's objective is to maximize the number of user clicks under the advertiser's budget while ensuring the revenue on each channel stays within an acceptable range. However, each channel's request is limited and cannot accommodate all advertisers. If all of them compete for the request on high-quality channels, it will certainly lift the bidding price and reduce advertisers' CPC satisfactory ratio. A well-designed planner may allocate different budgets for each channel based on the advertiser preference, as well as prevent budget over-allocation on specific channels.

A common solution is to leverage an advertiser-level constraint to limit the amount of budget allocation on the channel. It may result in reducing overall revenue since the bidding abilities of advertisers are quite different and thus one cannot allocate equal budget to all of them. Meanwhile, our design also faces a key challenge as how to restrict the updated policy to satisfy the channel-capacity constraint because it is hard to design a suitable reward function. Inspired by \cite{crossdqn}, we design a batch loss for the high-level planner to ensure that the budget allocated to each channel fluctuates within an acceptable range. Consider that we sample a batch of experiences containing multiple tuples $(s_p^h,a_p^h,r_p^h,c_p^h,s_{p+1}^h)$ from the high-level dataset $\mathcal{D}^h$, then the batch loss is calculated by:
\begin{equation}
\small
\label{loss-batch}
     \mathcal{L}_{Batch}(\theta) = \sum_{p=1}^P \bigg(\sum_{s_p^h} Cost\big(\mathop{\arg \max}_{a_p^h\in \mathcal{A}^h}Q_\theta^h(s_p^h,a_p^h)\big)-\kappa_p\bigg)^2,
\end{equation}
where $\kappa_p$ is calculated by multiplying historical CTR, CPC and impression count based on all advertisers in the batch, and a weight $w_p$ is introduced on the batch loss for learning stability.
Since the argmax function is not differentiable, a soft version is applied: 
\begin{equation}
\footnotesize
    Cost\big(\mathop{\arg \max}_{a_p^h\in \mathcal{A}^h}Q_\theta^h(s_p^h,a_{p}^h)\big)	\approx
      \sum_{n=1}^{|\mathcal{A}^h|}\frac{1}{Z}\exp \left[\beta Q_\theta^h(s_p^h,a_{p,n}^h)Cost(a_{p,n}^h)\right],
\end{equation}
where $\beta$ is the temperature coefficient and $Z=\sum_{n=1}^{|\mathcal{A}^h|}\exp [\beta Q^h_\theta(s_p^h,a_{p,n}^h)]$ is the normalization factor. While the high-level planner updates its strategy using randomly sampled batch experiences to maximize the number of clicks, the batch loss encourages to reallocate budgets across different channels based on historical statistics within the batch experiences. For example, it prompts some high-impact advertisers (i.e., those with higher conversion rates) to reduce their budget allocation on high-quality channels, thereby allowing low-impact advertisers to benefit from the superior advertising effects that these high-quality channels provide. This approach ensures that revenue on the channels does not fluctuate dramatically. Additionally, it prevents the high-level policy from consistently favoring higher investments in high-quality channels, which could lead to detrimental competition.


\subsection{Offline Data Augmentation by $\lambda$-Generation and Optimal $\lambda$-Selection for Online Adaptive Bidding}
\label{sec-lambda-generation}
As the advertiser's total budget and the platform's allocated budget may change over time, we need an adaptive bidding strategy that can dynamically respond to the budget changes. However, under the offline DRL training paradigm, the low-level executor learns the bidding strategy from the low-level dataset $\mathcal{D}^l$ which cannot be generalized to unseen budget cases, reducing the effectiveness of the high-level budget allocation. To this end, we adopt ``$\lambda$-generalization'' \cite{bcorle} to learn an adaptive bidding strategy for dynamically changing budgets.

In order to learn a bidding strategy that satisfies the budget constraint on each channel individually, the problem of c$^3$-bidding without CPC constraint can be converted into its Lagrangian dual problem \cite{lagranian} as:
\begin{equation}
\small
\begin{aligned}
    &\min_{\lambda}\max_{a^l_{i}}\sum_{i=1}^I Click(a^l_i)-  
    \lambda\bigg(\sum_{i=1}^ICost(a^l_i)-a_{p}^h\bigg) \\
    \Rightarrow&\min_{\lambda}\max_{a^l_{i}}\sum_{i=1}^I\bigg(Click(a_i^l)- 
    \lambda Cost(a_i^l)\bigg) +\lambda a_{p}^h\quad s.t.\quad \lambda \geq 0, 
\end{aligned} 
\end{equation}
where $\lambda$ is the Lagrangian multiplier that controls how much the bidding strategy spends. Thus, we take it as part of the input to the Q-network and modify the low-level reward function by:
\begin{equation}
\label{reward-modified}
    r^{l,\lambda}_i = r_i^l- \lambda c_i^l.
\end{equation}
Then the low-level policy can be formulated as:
\begin{equation}
\label{origin-pi}
    \pi^l(s_i^l,\lambda) = \mathop{\arg \max}_{a_i^l\in{\mathcal{A}^l}}Q_\theta^l(s_i^l,a_i^l,\lambda).
\end{equation}
Therefore, the low-level executor can adaptively respond to changing budgets by selecting the optimal $\lambda^*$. Next, we need to deal with the training of the corresponding policy under different $\lambda$ offline, and getting an optimal $\lambda^*$ under the budget constraint during online prediction.

During the offline training, we perform data augmentation by $\lambda$-generation, allowing the policy to learn how to bid under different $\lambda$. Given a fixed training dataset consisted of multiple tuples, we extend it by enlarging each tuple $(s_i^l,a_i^l,r_i^l,c_i^l,s_{i+1}^l)$ into $\{(s_i^l,a_i^l,r_i^{l,\lambda_n},c_i^l,s_{i+1}^l,\lambda_i^n)\}_{n=1}^N$, where $N$ is data repetition times and $\lambda_n$ is a uniformly sampled value from $[0,\lambda_{\max}]$. The range $[0,\lambda_{\max}]$ can guarantee that the cost of the learned policy falls within a controllable range \cite{bcorle}.
We also construct two additional evaluation networks $\hat{Q}^c_{\phi}$ and $\hat{Q}^u_{\eta}$ to accurately evaluate the expected cost and number of clicks under the low-level policy $\pi^l$. We use the action from $\pi^l$ instead of $\max$ operator when computing the target value in two evaluation network updates by:
\begin{equation}
\footnotesize
\begin{aligned}
\label{loss-eval}
    \mathcal{L}_{\hat{Q}}(\phi)=\mathbb{E}_{\tau^l\sim\mathcal{D}^l} \left[c_i^l+ \gamma^l\hat{Q}^c_{\phi}\big(s_{i+1}^l,\pi^l(s_{i+1}^l,\lambda_i^n),\lambda_i^n\big)
    -\hat{Q}^c_{\phi}(s_i^l,a_i^l,\lambda_i^n)\right],
\end{aligned}
\end{equation}

where $\tau^l$ is sampled trajectories from $\mathcal{D}^l$ and $c_i^l$ denotes the budget consumption under action $a_i^l$. Note that we utilize the same loss function $\mathcal{L}_{\hat{Q}}(\eta)$ for $\hat{Q}^u_{\eta}$ but compute number of clicks as $c_i^l$.

During the online prediction, two evaluation networks are used to perform $\lambda$-selection to ensure that the low-level policy does not exceed the budget allocated by the high-level planner. We uniformly sample $\{\lambda^n\}_{n=1}^{N_p}$ for each request $i$ within the range $[0,\lambda_{\max}]$, and select the $\lambda_i^*$ which satisfies the allocated budget $a_p^h$ as well as maximizing the number of clicks:
\begin{equation}
\label{lambda-selection}
    \lambda_i^* = \mathop{\arg \max}_{\lambda\in\{\hat{Q}^c_{\phi}(s_i^l,\pi(s_i^l,\lambda^n),\lambda^n)\leq a_p^h|_{n=1}^{N_p}\}} \hat{Q}^u_{\eta}(s_i^l,\pi^l(s_i^l,\lambda),\lambda), \forall i.
\end{equation}
In this way, the low-level executor can adaptively respond to the allocated budget $a_p^h$ and ensure the effectiveness of the budget allocation made by the higher-level planner.

 \subsection{CPC-guided Action Selection for Cross-channel Constraint Satisfaction }
 \label{sec-action-selection}
Due to the varying target users and advertising quality, the competition situation differs among channels, resulting in a significant discrepancy in terms of CPC between channels (i.e., high-quality channels have higher CPC). Therefore, we cannot simplify the cross-channel CPC constraint by setting the same target CPC for each channel. When we use Lagrangian relaxation to deal with both budget and CPC constraints simultaneously, it is impossible to find an effective pair of Lagrangian multipliers to satisfy them due to the explosion of the solution space. We design a CPC-guided action selection (CPC-AS) scheme to help the low-level executor choose the action that satisfies the CPC constraint by considering both the past and the future.

When making a decision for a request $i$, the final CPC for an advertiser $m$ is divided into two parts:
\begin{equation}
\begin{aligned}
CPC_m^{real} &=\frac{\sum_{p=1}^{P}\sum_{i=1}^{I} Cost(a^l_{p,i})}{\sum_{p=1}^{P}\sum_{i=1}^{I} Click(a^l_{p,i})} 
\\ &=\frac{Cost_{m,t}+\sum_{p=1}^P\sum_{i=t}^{I} Cost(a^l_{p,i})}{Click_{m,t}+\sum_{p=1}^P\sum_{i=t}^{I} Click(a^l_{p,i})},
\end{aligned}
\end{equation}

where $Cost_{m,t}$ and $Click_{m,t}$ denote the costs and number of clicks already happened up to now, respectively. 
As we describe in Section \ref{sec-lambda-generation}, two evaluation networks $\hat{Q}^c_{\hat{\theta}}$ and $\hat{Q}^u_{\hat{\theta}}$ are developed to estimate the expected discounted costs and the number of clicks, respectively. Given the current state-action pair $(s^l_{i},a^l_{i})$, we approximate the expected costs and number of clicks through $\hat{Q}^c_{\phi}(s^l_{i},a^l_{i},\lambda)$ and $\hat{Q}^u_{\eta}(s^l_{i},a^l_{i},\lambda)$. Therefore, we define $CPC_m^{pred}(a_{i}^l)$ by combining the past and expected future by:
\begin{equation}
\begin{aligned}
CPC_m^{pred}(a_{i}^l) = \frac{Cost_{m,t}+\sum_{p=1}^P\sum_{i=t}^{I}(\gamma^l)^{i-1}Cost(a^l_{p,i})}{Click_{m,t}+\sum_{p=1}^P\sum_{i=t}^{I}(\gamma^l)^{i-1}Click(a^l_{p,i})}\\
= \frac{Cost_{m,t}+\sum_{p=1}^P\hat{Q}^c_{\phi}(s^l_{p,i},a^l_{p,i},\lambda)}
{Click_{m,t}+\sum_{p=1}^P\hat{Q}^u_{\eta}(s^l_{p,i},a^l_{p,i},\lambda)}. \label{CPC-calculate}
\end{aligned}
\end{equation}
Since ad requests from all channels arrive in sequence, and each request only belongs to one channel by definition, we cannot estimate the future of other channels, and then we save their most recent estimations as the input.
Together with $CPC(a_i^l)$, the low-level policy becomes:
\begin{equation}
\label{mask-policy}
    \pi^l(s^l_i,\lambda) = \mathop{\arg \max}_{a\in\{{CPC_m^{pred}(a_i^l)\leq CPC_m^{set}|a_i^l\in\mathcal{A}^l\}}}Q^l_\theta(s^l_i,a,\lambda).
\end{equation}
Note that there exists a slight bias between $CPC_m^{real}$ and $CPC_m^{pred}(a_{i}^l)$ when $\gamma^l< 1$. In Section \ref{proof}, we experimentally prove that the bias becomes smaller when $\gamma^l$ is close to $1$, which can be ignored in practice.

\begin{algorithm}[tb]
\caption{Offline Training}
\label{algo_train}
  \KwIn{
  Log data $\mathcal{D}$, high-level CVAE $G^h_\omega$, Q-network $Q^h_\theta$ and target network $Q^h_{\theta'}$, Low-level CVAE $G^l_\omega$, Q-network $Q^l_\theta$ and target network $Q^l_{\theta'}$, evaluation networks $\hat{Q}^u_{\eta}$ and $\hat{Q}^c_{\phi}$.}
   Initialize all parameterized networks;\\
   Process log data into high-level dataset $\mathcal{D}^h$ and low-level dataset $\mathcal{D}^l$;\\
  \For{High-level update iteration$=1,2,3,\dots$}{
    Sample a batch contains $J^h$ tuples $\{(s_j^h,a_j^h,r_j^h,c_j^h,s_{j+1}^h)\}_{j=1}^{J^h}$ from $\mathcal{D}^h$;\\
    Update high-level CVAE by minimizing Eqn. (\ref{loss-vae});\\
    Calculate the $\mathcal{L}_Q(\theta)$, $\mathcal{L}_{OOD}(\theta)$  and the batch loss $\mathcal{L}_{Batch}(\theta)$ by Eqn. (\ref{loss-q}), (\ref{loss-ood}) and (\ref{loss-batch}), .\\
    Update Q-network $Q_\theta^h$ by minimizing the weighted loss $w_1\mathcal{L}_Q(\theta)+w_2\mathcal{L}_{OOD}(\theta)+w_b\mathcal{L}_{Batch}(\theta)$;\\
    Every $N_{target}$ iterations synchronize $Q^h_{\theta'} \xleftarrow{} Q^h_{\theta}$;
    }
  \For{Low-level update iteration$=1,2,3,\dots$}{
    Sample a batch of experiences contains $J^l$ tuples $\{(s_j^l,a_j^l,r_j^l,c_j^l,s_{j+1}^l)\}_{j=1}^{J^l}$ from $\mathcal{D}^l$;\\
    Update low-level CVAE by minimizing Eqn. (\ref{loss-vae});\\
    \For{$j=1,\dots,J^l$}{
    Get the allocated budget $a^h_p$ using $Q^h_\theta$;\\
    Uniform sample $\{\lambda^n\}_{n=1}^N$ in range $[0,\lambda_{\max}]$;\\
    Calculate the reward $r_j^{l,\lambda_j^n}$ by Eqn. (\ref{reward-modified});\\ 
    Including $\lambda^n$ and $a^h_p$ into the $s_j^l$, then enlarging each tuple into $\{(s_j^l,a_j^l,r_j^{l,\lambda^n},c_j^l,s_{j+1}^l,\lambda_j^n)\}_{n=1}^N$;
    }
    Calculate the $\mathcal{L}_Q(\theta)$ and $\mathcal{L}_{OOD}(\theta)$ by Eqn. (\ref{loss-q}) and (\ref{loss-ood}).\\
    Update Q-network $Q_\theta^l$ by minimizing the weighted loss $w_1\mathcal{L}_Q(\theta)+w_2\mathcal{L}_{OOD}(\theta)$;\\
    Update evaluation networks $\hat{Q}^u_{\eta}$ and $\hat{Q}^c_{\phi}$ by minimizing the $\mathcal{L}_{\hat{Q}}(\eta)$ and $\mathcal{L}_{\hat{Q}}(\phi)$ in Eqn. (\ref{loss-eval});\\
    Every $N_{target}$ iterations synchronize $Q^l_{\theta'} \xleftarrow{} Q^l_{\theta}$;
    }
\end{algorithm}

\begin{algorithm}[tb]
\caption{Online Prediction}
\label{algo_pred}
  \KwIn{Trained high-level Q-network $Q_\theta^h$, low-level Q-network $Q_\theta^l$, and evaluation networks $\hat{Q}^u_{\eta}$ and $\hat{Q}^c_{\phi}$.}
\While{\text{Incoming ad request} $i$}{
    Get advertiser-level feature $s_p^h$ and request-level feature $s_i^l$ from the platform. \\
    Allocate budget $a^h_p$ by $\mathop{\arg \max}_{a_p^h\in{\mathcal{A}^h}}Q_\theta^h(s_p^h,a_p^h).$ \\
    Find the optimal $\lambda_i^*$ according to $a_p^h$ by Eqn. (\ref{lambda-selection});\\
    Calculate $\{CPC_m^{pred}(a_i^l)|a_i^l\in\mathcal{A}^l\}$ by Eqn. (\ref{CPC-calculate})\\
    Calculate bidding action $a_i^l$ by Eqn. (\ref{mask-policy})
}

\end{algorithm}


\subsection{Algorithm Description}
\subsubsection{Offline Training} 
We first show the pseudo-code for offline training in Algorithm $1$. At the beginning, we initialize all parameterized networks (Line $1$). Then we process the data for the high-level and low-level training individually (Line $2$), to use the available log data efficiently. For the high-level planner,
after sampling a batch of experiences from $\mathcal{D}^h$ (Line $4$), we update CVAE first by Eqn. (\ref{loss-vae}) (Line $5$). Using sampled experiences, we calculate the $\mathcal{L}_Q(\theta)$ and $\mathcal{L}_{OOD}(\theta)$ as well as $L_{Batch}(\theta)$ by Eqn. (\ref{loss-q}), (\ref{loss-ood}) and (\ref{loss-batch}), then update high-level Q-network with weighted loss (Line $6$-$7$). Finally, the target network $\theta'$ is synchronized periodically (Line $8$).
For the low-level executor, we sample a batch of experiences from $\mathcal{D}^l$ and then update the low-level CVAE by Eqn. (\ref{loss-vae}) (Line $10$-$11$). With each sampled experience, we leverage the trained high-level Q-network to determine the allocated budget (Line $13$). Then, we augment the origin tuple by sampling multiple $\{\lambda^n\}_{n=1}^N$ and modifying the reward (Line $14$-$15$). The sampled $\lambda^n$ and allocated budget $a^h_p$ are incorporated into the state $s_j^l$ (Line $16$). Using the augmented tuple, we update the low-level Q-network with weighted loss by Eqn. (\ref{loss-q}) and (\ref{loss-ood}) (Line $17$-$18$). Two evaluation networks are updated by minimizing $\mathcal{L}_{\hat{Q}}(\eta)$ and $\mathcal{L}_{\hat{Q}}(\phi)$ in Eqn. (\ref{loss-eval}) (Line $19$). 

\subsubsection{Online Prediction}
The pseudo-code for online prediction is given in Algorithms 2. For each ad request $i$, 
HiBid gets advertiser-level feature $s_p^h$ and request-level feature $s_i^l$ as in Section $3$ from the advertising platform (Line $2$). Then, the high-level planner allocates budget $a_p^h$ by taking advertiser-level information as the input of $Q_\theta^h$. (Line $3$) Together with two evaluation networks $\hat{Q}^u_{\eta}$ and $\hat{Q}^c_{\phi}$, the low-level executor finds the optimal $\lambda_i^*$ according to Eqn. (\ref{lambda-selection}) (Line $4$). To satisfy the CPC constraint, $CPC_m^{pred}(a_i^l)$ for each action is calculated by Eqn. (\ref{CPC-calculate}) (Line $5$). Considering both the output of Q-network $Q_\theta^l(s_i^l,\lambda_i^*)$ and $\{CPC_m^{pred}(a_i^l)|a_i^l\in\mathcal{A}^l\}$, the low-level executor gives the final bidding action based on Eqn. (\ref{mask-policy}) (Line $6$).


\section{Experiment}
\label{sec-exp}
\subsection{Setup}
\subsubsection{Dataset}We use large-scale log data collected from Meituan advertising system (which is physically deployed online and running for real-time services) for offline training and performance evaluation. The data contains $28$ days of bidding logs and is divided into two parts, i.e., $21$ days for training and $7$ days for evaluation. On average, we sampled $64,272$ advertisers for $70$ million ad requests on $4$ channels in a day. We follow the common setting in \cite{mcq} for most hyper-parameters (leaving two key ones for hyper-parameters tunning in Section \ref{sec-hyper-tunning}) and list them in Table \ref{table_hyper}. 

\subsubsection{Offline Evaluation System} Deploying a model to an online system without evaluating its potential effects is risky, since it may lead to significant loss of revenue.
To avoid this, we design an offline evaluation system for cross-channel bidding scenarios, which includes two modules: advertising system simulator and user feedback predictor. The former simulates Meituan's real online advertising platform, including the process of retrieval, bidding, ranking and pricing. The latter is used to predict the user's feedback on certain advertisements and provide the advertising results for evaluation.

\subsubsection{Evaluation Metrics}
We introduce five metrics to mathematically evaluate the performance of HiBid in the considered c$^3$-bidding problem, including (a) total impression counts ($Impr$), (b) total number of clicks ($Click$), (c) average CPC ($CPC$), (d) average CPC satisfactory ratio ($CSR$), and (e) average ROI ($ROI$) of all advertisers. In particular, $CSR$ is average of $\mathds{1}_{CPC_m^{real}\leq CPC_m^{set}}$ of all advertisers and $\mathds{1}$ is the indicator function.
To accurately evaluate the effectiveness of various methods and eliminate the influence of specific application scenarios, we use the normalized score with respect to the statistical result obtained from the log data (offline performance evaluation), and online solution R-BCQ \cite{bcorle} (online A/B testing) for each metric.

\subsubsection{System Setup}
We implement Hibid with Tensorflow 1.15 and utilize $4$ NVIDIA A100 GPUs for offline training. The training time cost is directly proportional to the choice of $N$, and it lasts about $29$ hours when $N=30$. For online prediction, we deployed Hibid on $233$ servers, each of which is equipped with an Intel(R) Xeon(R) Platinum 8352Y CPU @ 2.20GHz and an NVIDIA A30 GPU. The maximum number of concurrent requests is about $18,899$ (during the business peak period) and the inference time is shown in Table \ref{table_online}.

\begin{table}[tb]
\centering
\linespread{2}
\caption{Key hyper-parameters in HiBid}
\begin{tabular}{cc}
\cline{1-2}
Hyper-parameter & Value   \\ \cline{1-2}
high-level and low-level batch sizes   &     4096, 1024                \\
high-level and low-level learning rates   &  1e-5, 1e-5      \\
discounted factors $\gamma^h,\gamma^l$     &  $1, 0.999$      \\
loss weights  $w_1,w_2,w_b$   &    $1,0.05,0.1$    \\
high-level decision interval & 1 day \\
$\lambda$ sampling range $\lambda_{\max}$ & 1.45 \\ 
data repetition times  $N$   &     $30$   \\

number of sampled $\lambda$ in online prediction $N_p$   &     $50$   \\
low level action range $[a^l_{\min},a^l_{\max}]$ & $[0.5,1.5]$ \\
\cline{1-2}
\end{tabular}
\label{table_hyper}
\end{table}

\subsection{Hyper-parameter Tuning}
\label{sec-hyper-tunning}
We first show the results of hyper-parameters tunning in HiBid, as loss weight $w_b$ (in batch loss; see Section \ref{sec-batch-loss}) and data repetition times $N$ (in $\lambda$-generalization; Section \ref{sec-lambda-generation}). We tune $w_b \in \{0,0.02,0.04,\dots,0.2\}$ to investigate the effect of weighted batch loss on budget allocation and $N \in \{0,1,5,10,20,30,40,50,60\}$ to study the impact of sample efficiency on bidding strategy. 

\begin{figure}[tb]
\centering
\begin{subfloat}[Impact of $w_b$.]{
\resizebox{0.8\columnwidth}{0.6\columnwidth}{
    \centering
    \includegraphics[width=\columnwidth]{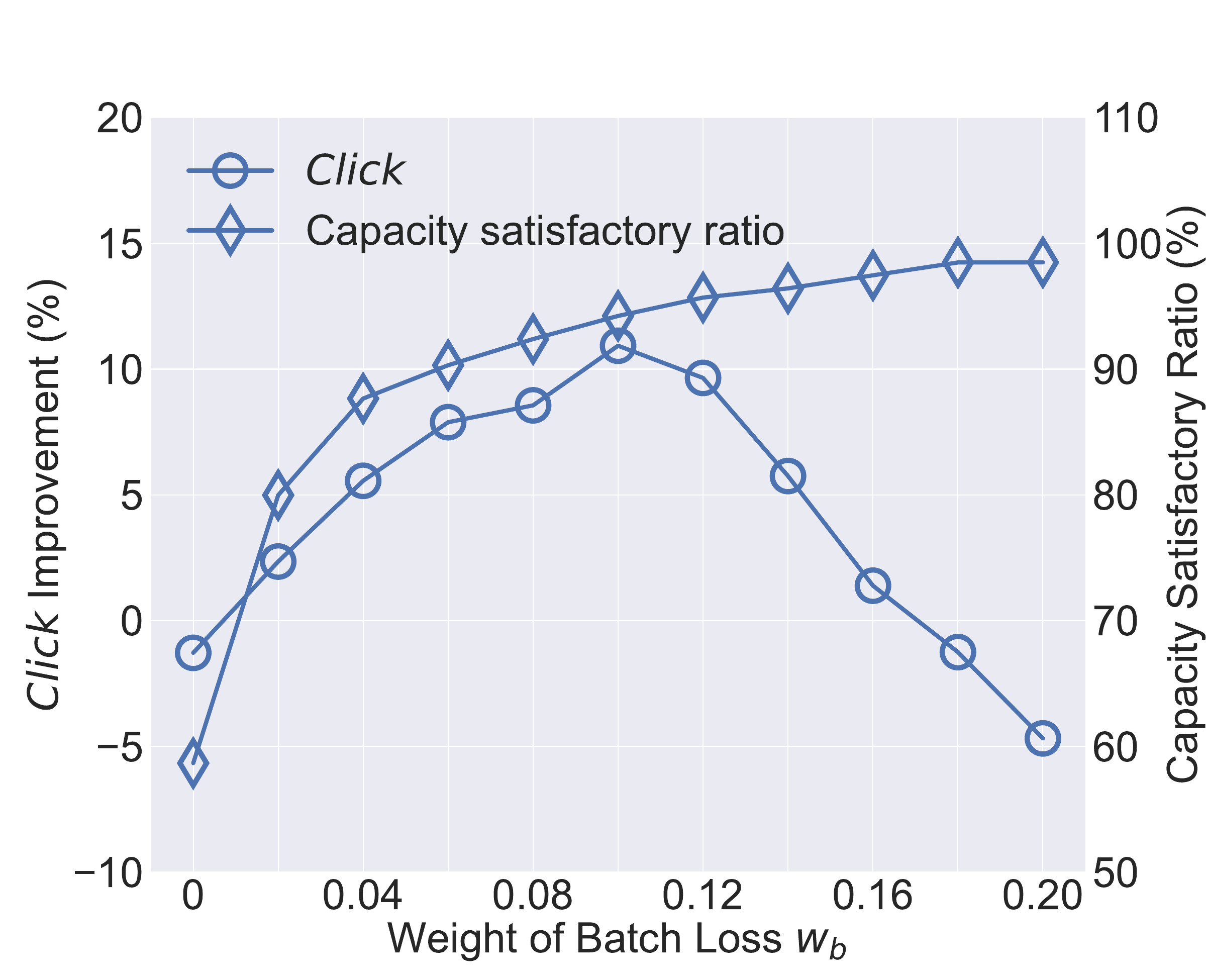}
    \label{fig-hyper-batchloss}
    \vspace{-6mm}
 }
}
\end{subfloat}
\begin{subfloat}[Impact of $N$.]{
\resizebox{0.8\columnwidth}{0.6\columnwidth}{
    \centering
    \includegraphics[width=\columnwidth]{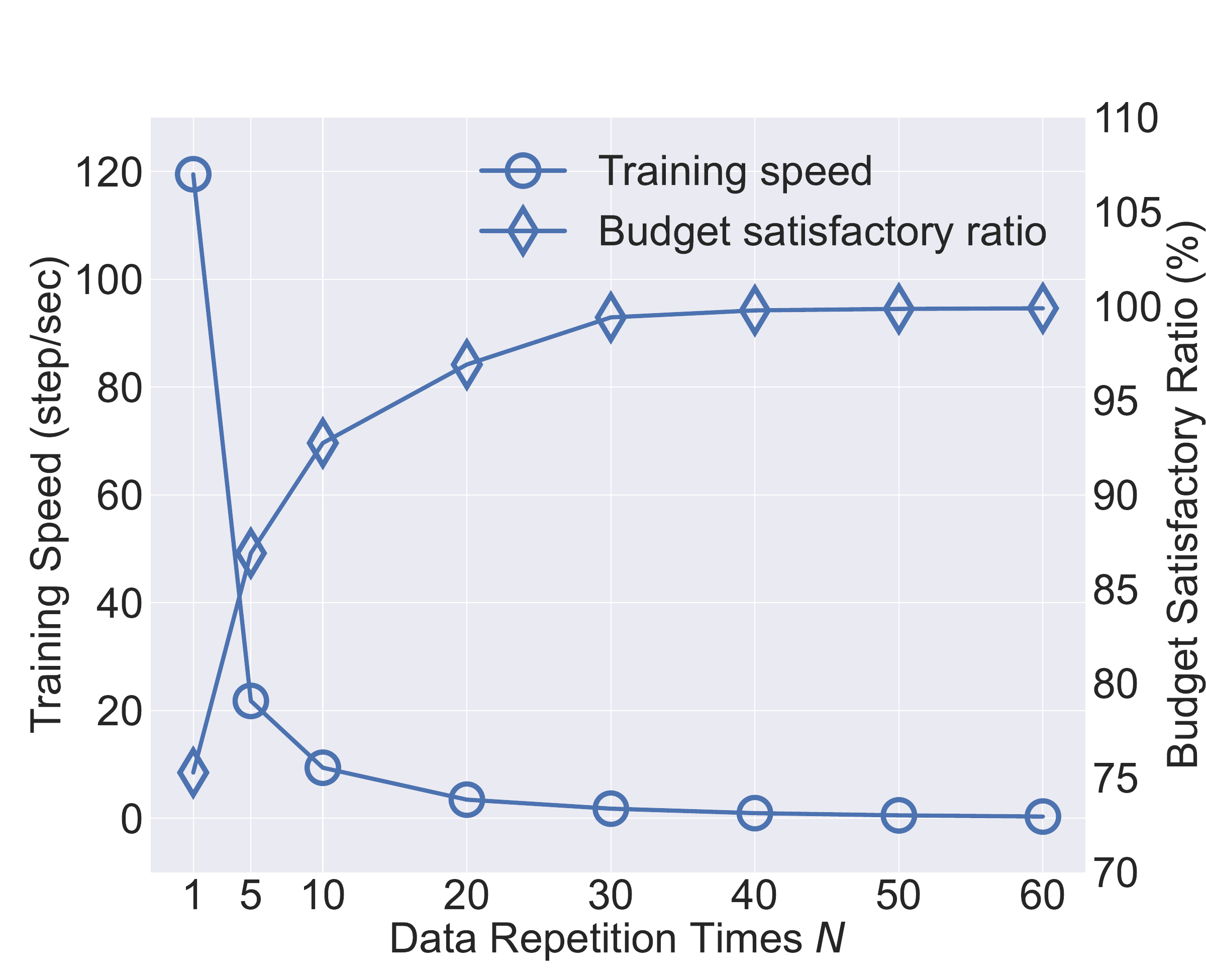}
        \vspace{-3mm}
    \label{fig-hyper-lambda}
 }
 }
\end{subfloat}
\caption{Hyper-Parameters Tunning. The Capacity satisfactory ratio is calculated as the average of $\mathds{1}_{\left|\sum_{m=1}^M  Cost(a^h_{m,p}) - \kappa_p \right| \leq \epsilon}$ of all channels and Budget satisfactory ratio is calculated by the average of $\mathds{1}_{\sum_{i=1}^I Cost(a^l_{p,i})\leq a^h_{p}}$ of all advertisers, respectively.
}
\label{fig-hyper}
\end{figure}

During practical deployment, the platform requirement is that the fluctuation of revenue does not exceed $1\%$. Thus, $\epsilon$ is set to $0.01\kappa_p$ for each channel $p$. From Figure \ref{fig-hyper}(a), we see that the capacity satisfactory ratio increases when we increase the $w_b$. However, the \textit{Click} reaches the peak when $w_b=0.1$ and then gradually decreases. This is because updating the budget allocation strategy with a large $w_b$ can prevent the over-allocation issue on high-quality channels. When $w_b$ is too large, the batch loss will have a negative impact on the policy updating process, resulting in a poor budget allocation strategy and advertising results. As shown in Figure \ref{fig-hyper}(b), we observe that as the data repetition times $N$ increases, the lower-level executor is able to accurately satisfy the allocated budget since more experiences help the bidding strategy generalize to unseen budget cases. However, overuse of the training experiences may result in poor training efficiency. Therefore, batch loss weight $w_b=0.1$ and data repetition times $N=30$ are the two best hyper-parameters chosen for performance comparison hereafter.

\begin{table}[]
\setlength\tabcolsep{1pt}
\linespread{1.4}
\scriptsize
\centering
\caption{Ablation Study}
\begin{tabular}{cc|ccccc}
\hline
\multicolumn{2}{c|}{Method}                                                                    & \textit{Impr}   & \textit{Click}   & \textit{ROI}    & \textit{CPC}     & \textit{CSR}    \\ \hline
\multicolumn{2}{c|}{HiBid}                                                                     & \textbf{0.03\%} & \textbf{10.93\%} & \textbf{4.53\%} & \textbf{-6.14\%} & \textbf{8.94\%} \\ \hline
\multicolumn{1}{c|}{\multirow{2}{*}{\makecell[c]{High\\level}}} & .. w/o batch loss                           & -4.50\%         & -1.28\%          & 3.44\%          & -0.03\%          & 2.76\%          \\
\multicolumn{1}{c|}{}                            & .. w/o budget allocation                    & -5.93\%         & -1.97\%          & 4.32\%          & -0.88\%          & 2.02\%          \\ \hline
\multicolumn{1}{c|}{\multirow{3}{*}{\makecell[c]{Low\\level}}}  & .. w/o $\lambda$-generalization             & -3.35\%         & -1.14\%          & 5.21\%          & -1.71\%          & 4.67\%          \\
\multicolumn{1}{c|}{}                            & .. w/o CPC-AS                               & -2.70\%         & 3.99\%           & -2.88\%         & 6.47\%           & -6.54\%         \\
\multicolumn{1}{c|}{}                            & .. w/o   $\lambda$-generalization  \& CPC-AS & -6.44\%         & 8.43\%           & -7.90\%         & 12.26\%          & -9.87\%         \\ \hline
\end{tabular}
\label{table-ablation}
\vspace{-3mm}
\end{table}




 \begin{figure*}[tb]
\centering
\vspace{-3mm}
\subfloat[$Click$]{
  \includegraphics[width = 0.45\textwidth]{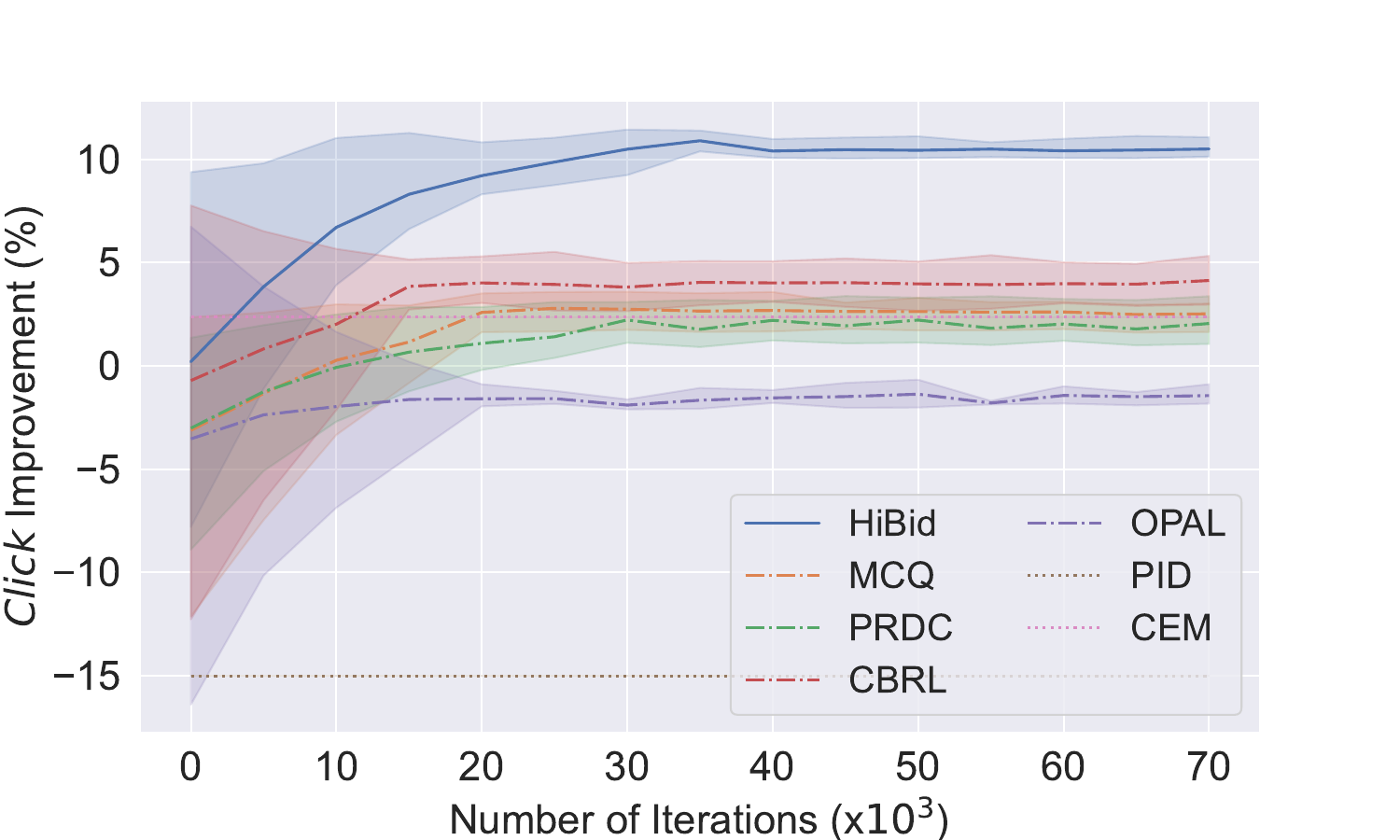}%
  \label{fig:sf_compare_click}
}
\subfloat[$CSR$]{
  \includegraphics[width = 0.45\textwidth]{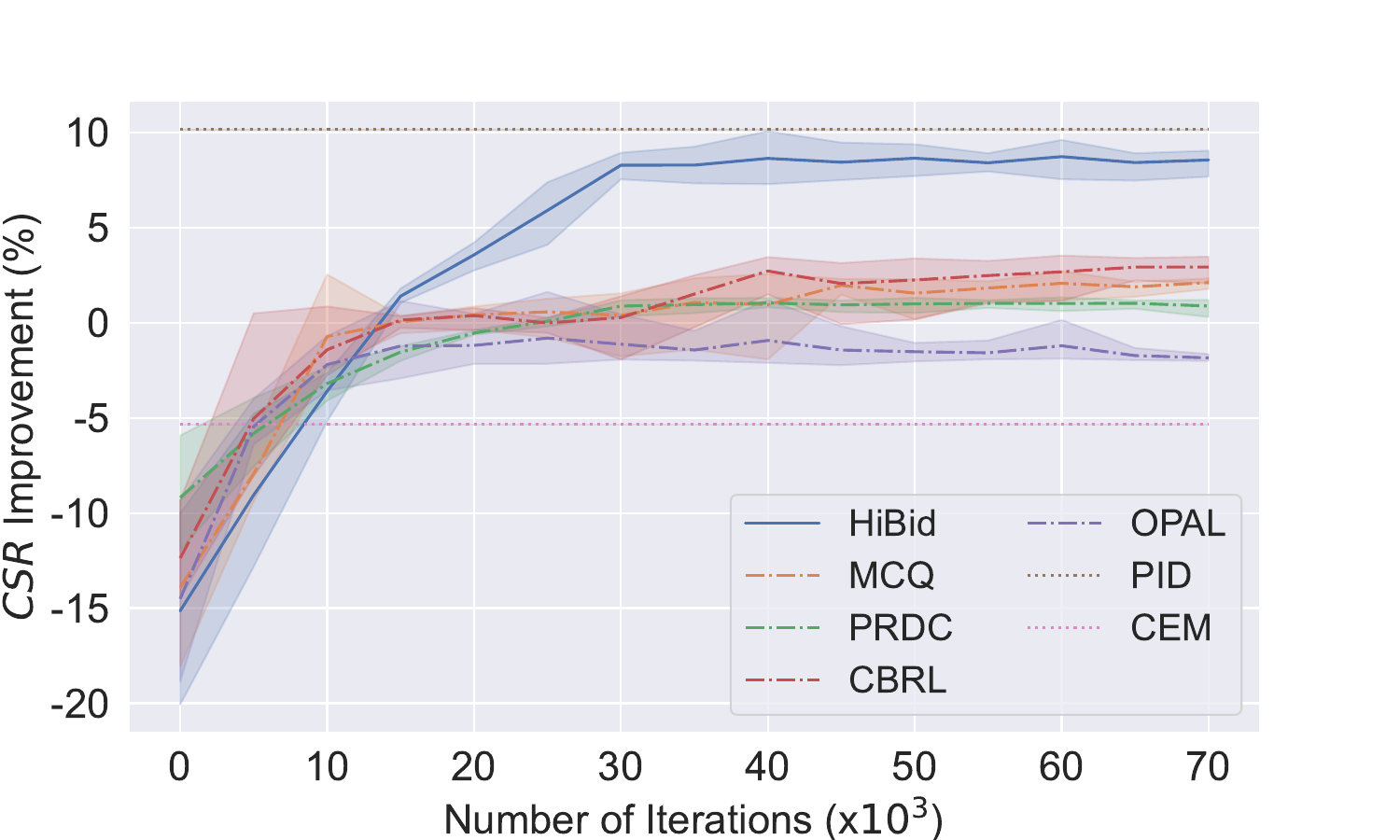}%
  \label{fig_compare_car}
}
\caption{Offline performance comparison with 5 baselines in terms of $Click$ and $CSR$ improvements.}
\label{fig_comparison_1}
\end{figure*}

 \begin{figure*}[tb]
\centering
\vspace{-5mm}
\subfloat[$ROI$]{
  \includegraphics[width = 0.33\textwidth]{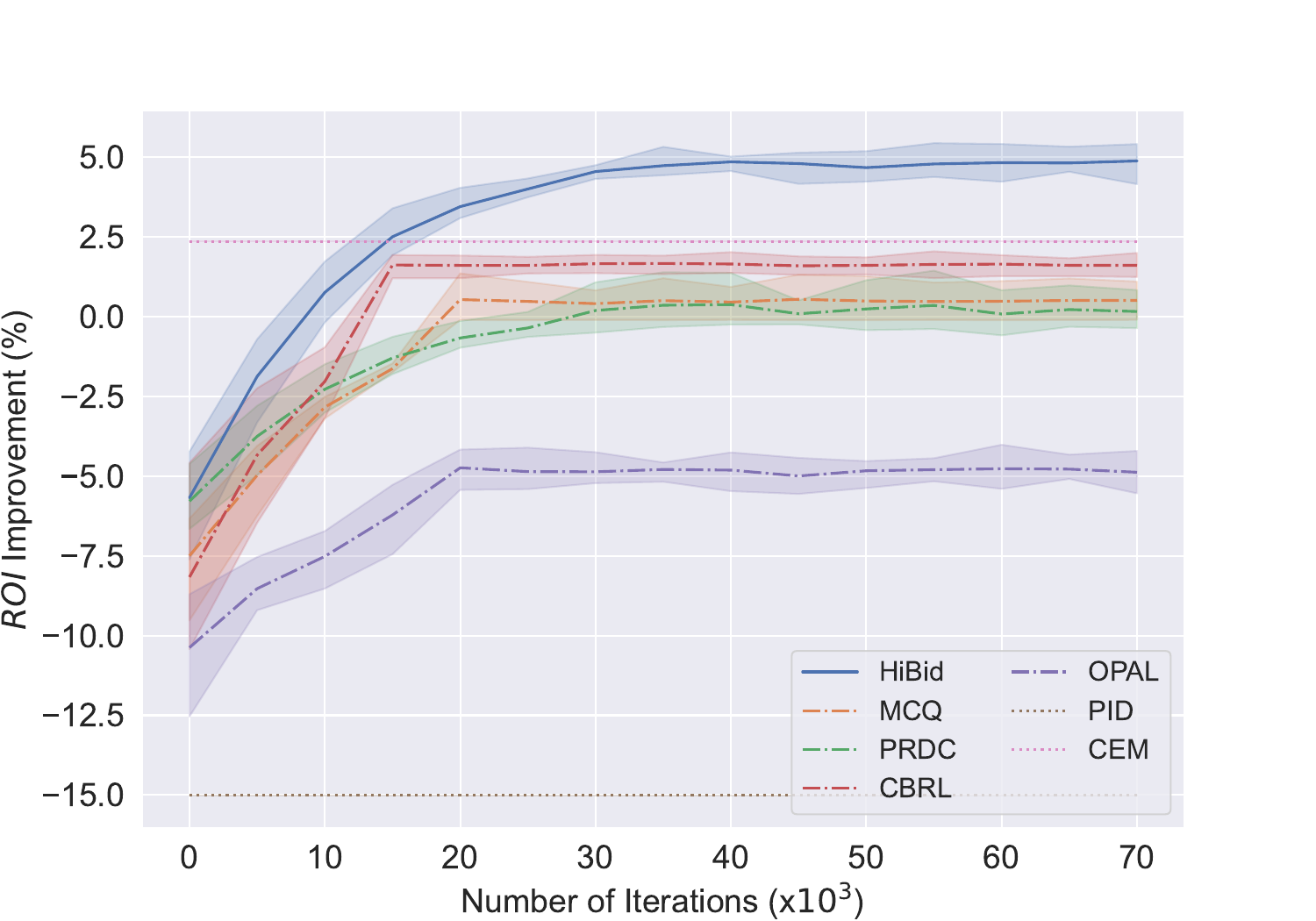}%
  \label{fig:sf_compare_roi}
}
\hspace{-5mm}
\subfloat[$Impr$]{
  \includegraphics[width = 0.33\textwidth]{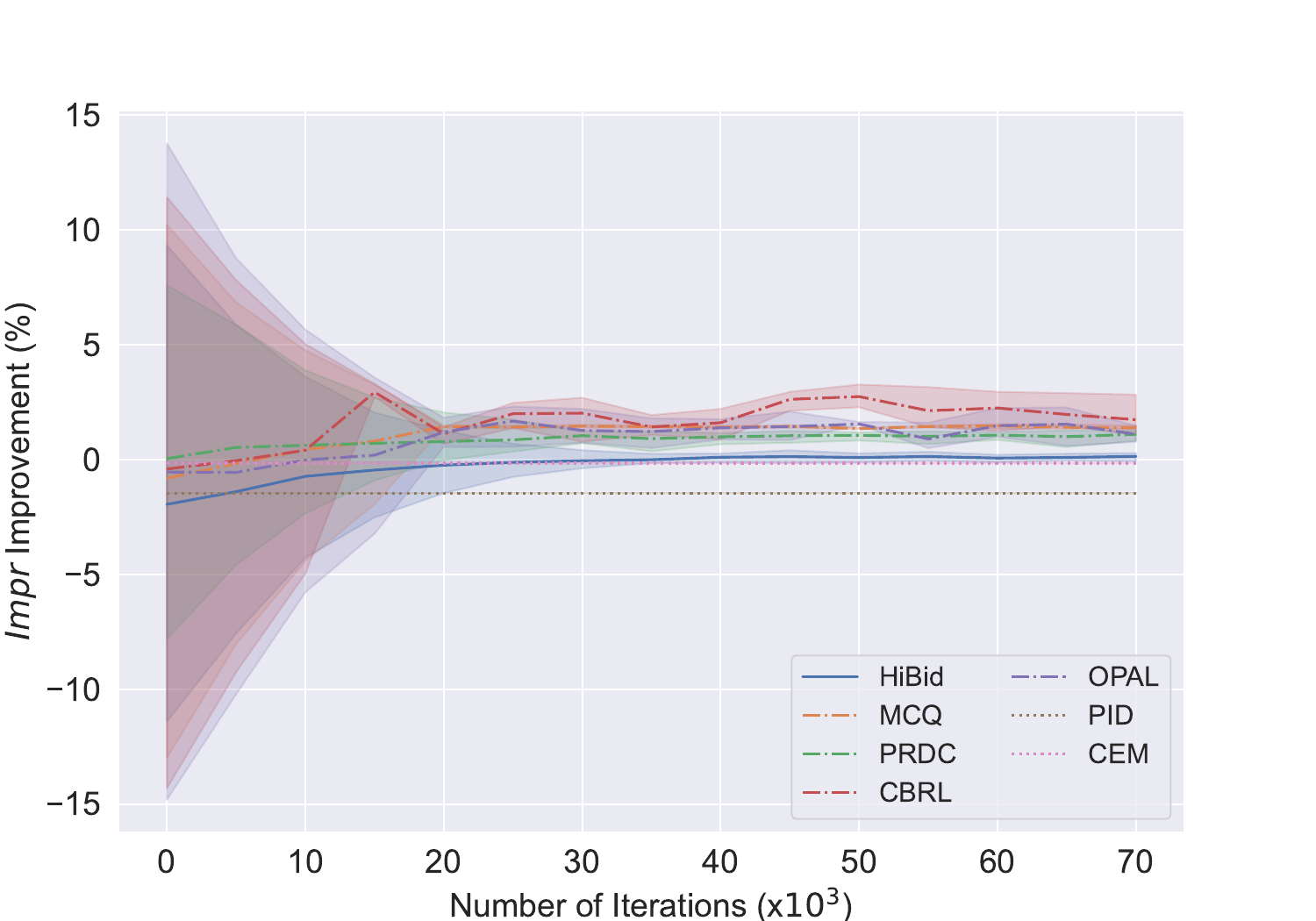}%
  \label{fig_compare_impr}
}
\hspace{-5mm}
\subfloat[$CPC$]{
  \includegraphics[width = 0.33\textwidth]{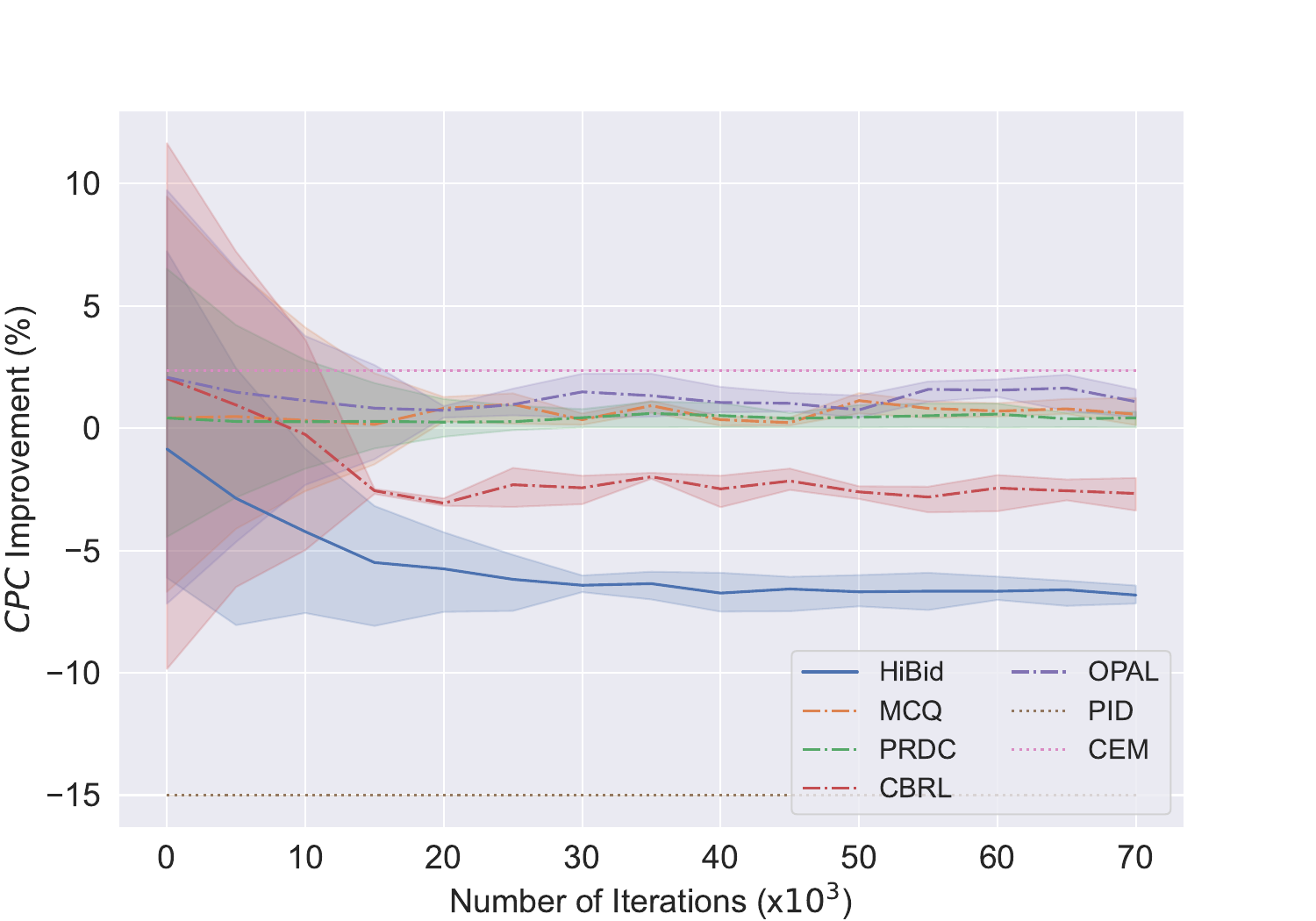}%
  \label{fig_compare_cpc}
}
\caption{Offline performance comparison with 5 baselines in terms of $ROI$, $Impr$ and $CPC$ improvements.}
\label{fig_comparison_2}
\end{figure*}
\subsection{Ablation Study}
We gradually remove four key components from the high-level planner and the low-level executor in HiBid, to verify the benefits brought by each component. The results are shown in Table \ref{table-ablation}.

We first show the benefits of batch loss and budget allocation in the high-level planner. When the batch loss module is removed, $CPC$ increases $6.11\%$, and $CSR$ significantly drops from $8.94\%$ to $2.76\%$. We also present the specific changes in $Impr$ and $CPC$ across four channels in Table \ref{table_batch_loss}. We observe that when we removed the batch loss module, due to limited supply from high-quality channels (i.e., Channel 1), advertisers exhibite a stronger investment intention (as an increased value in $Impr$), which results in increased advertising cost (higher CPC) on high-quality channels. However, the summed $Impr$ on four channels is decreased, and an overall increase in average $CPC$ is observed, which is unfavorable for either advertiser's investment or platform itself. 
$Click$ decreases from $10.93\%$ to $-1.97\%$ when we removed the entire budget allocation module. In this way, the low-level executor bids for each incoming request without a budget, leading unsuitable advertisers to take away and waste the click opportunities on that channel.

\begin{table}[]
\setlength\tabcolsep{1.2pt}
\linespread{1.5}
\footnotesize
\centering
\caption{Impact of batch loss}
\begin{tabular}{c|c|cccc}
\hline
Metric     & Method               & Channel 1 & Channel 2 & Channel 3 & Channel 4 \\ \hline
\multirow{2}{*}{$Impr$} & HiBid                & 0.01\%    & 0.03\%    & 0.03\%    & 0.04\%    \\
                      & .. w/o batch loss & 7.64\%    & -7.28\%   & -8.85\%   & 8.56\%    \\ \hline
\multirow{2}{*}{$CPC$}  & HiBid                & -3.65\%   & -5.88\%   & -7.29\%   & -10.86\%  \\
                      & .. w/o batch loss & 6.76\%    & -1.67\%   & -2.24\%   & -1.65\%   \\ \hline
\end{tabular}
\label{table_batch_loss}
\end{table}

We further observe the impact of $\lambda$-generalization and CPC-AS on the low-level bidding strategy. Compared to HiBid without $\lambda$-generalization, $Click$ and $CSR$ of HiBid achieves $12.07\%$ and $4.27\%$ improvements, respectively. This is because the low-level executor accurately adapts its bidding strategy according to the allocated budget, and avoids taking away the budget originally from other channels, bringing an improvement in advertising performance. When we removed CPC-AS, $CSR$ drops significantly from $8.94\%$ to $-6.54\%$ since the higher bids result in more clicks but exhibit a negative impact on the advertisers' expectations. When both $\lambda$-generalization and CPC-AS are removed, $Impr$ and $ROI$ drastically drop, which confirms the benefits of introducing $\lambda$-generalization and CPC-AS as the contribution of this paper.




\begin{figure*}[]
\centering
\includegraphics[width=0.85\textwidth]{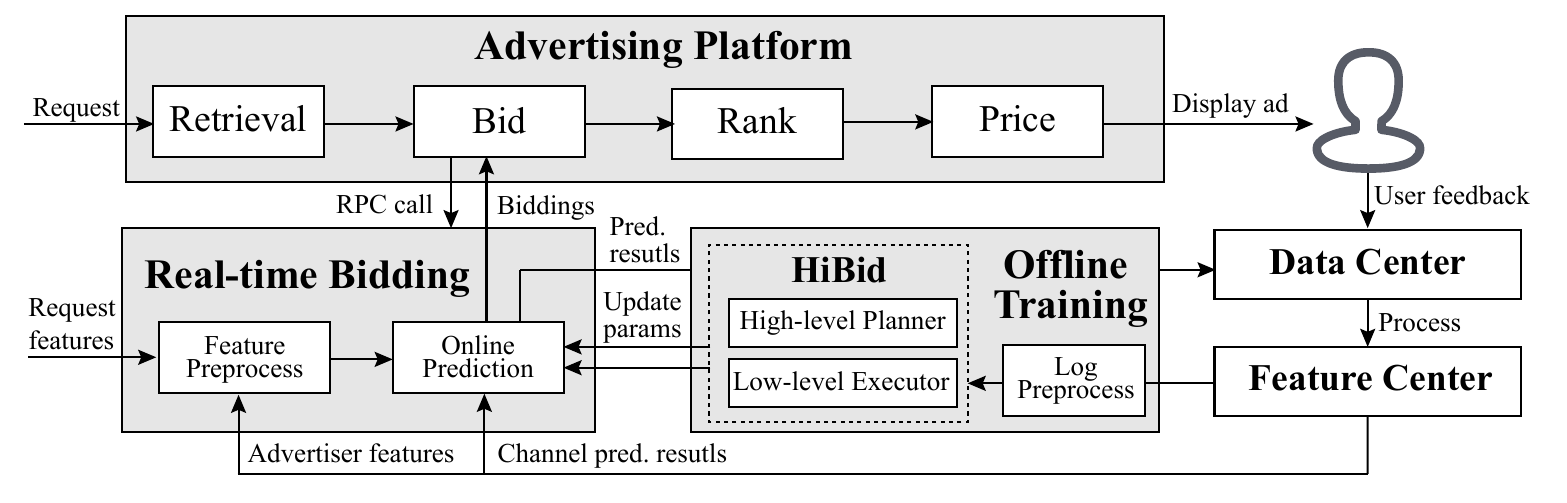}
\caption{HiBid deployment in Meituan advertising platform.}
\label{fig_deployment}
\end{figure*}

\subsection{Offline Performance Comparison}
We compare HiBid with six baselines:
\begin{enumerate}
    \item \textbf{CBRL} \cite{cbrl}: It is a curriculum-guided Bayesian reinforcement learning (CBRL) framework with an indicator-augmented reward function to adaptively control the constraint-objective trade-off for ROI-constrained single-channel bidding, which is considered as the state-of-the-art approach. For a fair comparison, we adapt CBRL to c$^3$-bidding problem by replacing ROI constraint with CPC constraint while maintaining the origin training process.
    \item \textbf{OPAL} \cite{opal}: Its high-level agent is trained by unsupervised learning, which provides a temporal abstraction for the low-level agent to improve offline policy optimization. It is considered as the state-of-the-art hierarchical offline DRL method. We adopt it by training a CVAE for budget allocation with the same network structure as HiBid.
    \item \textbf{MCQ} \cite{mcq}: It alleviates the value overestimation effect that occurred in OOD actions by actively training and correcting their Q values. We consider it as the state-of-the-art approach in offline DRL.
    \item \textbf{PRDC} \cite{PRDC}: It is another offline DRL method with a policy regularization mechanism. It employs dataset constraints to allow the policy to choose better actions that do not appear in the dataset with the nearest neighbor retrieval, while still maintaining sufficient conservatism for OOD actions.
    \item \textbf{CEM} \cite{cem}: Cross-Entropy Method (CEM) is a popular evolutionary algorithm, where we consider the c$^3$-bidding problem as a black-box optimization problem, aiming to maximize the number of clicks under the total budget and CPC constraints. Note that we always choose the policy set that is close to $CPC_m^{set}$ for policy updating in a CEM iteration.
    \item \textbf{PID} \cite{pid}: Proportional-Integral-Derivative (PID) controller is a classical feedback controller that is widely adopted, and performs well in unknown environments. We keep the advertiser's current CPC close to the target $CPC_m^{set}$ to satisfy the advertiser's cross-channel CPC constraint.
\end{enumerate}
Since OPAL, MCQ and PRDC do not meet the CPC and budget constraints, we explicitly consider these two by designing a reward function, that is able to maximize $Click$ from Meituan's past online service experiences, as:
\begin{align}
\big(1+w_x*\overline{CPC}*(1+CPC_m^{set})\big)*r_i^l-w_y*c^i_l,
\end{align}
where $\overline{CPC}$ is the calculated average CPC, $w_x=1.35$ and $w_y=1.31$ are manually set weights. Results are shown in Figure \ref{fig_comparison_1} and Figure \ref{fig_comparison_2}. 
By combining budget allocation and constrained bidding, HiBid enhances the matching efficacy between ads and users, thereby improving the $CPC$ and $CSR$ without taking extra impressions away from others. $Impr$ is influenced by the average bidding price decided by the proposed low-level executor since there are other advertising products competing for these ad requests. The slight fluctuation in HiBid's $Impr$ indicates that the average bidding price from HiBid is closer to the baseline. This is attributed to the proposed batch loss, which constrains the bidding expenditure of advertisers on different channels, as demonstrated in Table \ref{table_batch_loss}.
The second-best method is CBRL with $3.62\%$ and $3.24\%$ improvements over online solution R-BCQ in terms of $Click$ and $CSR$, respectively. CBRL gradually learns the bidding strategy that satisfies constraints through curriculum learning, but its performance is still worse than HiBid due to the lack of accurate budget allocation.
MCQ actively updates OOD state-action pairs using the experiences from the dataset, and the learned conservative strategy might miss out on some high-revenue actions. Therefore, as the number of model training steps increases, $CSR$ slightly improves, but $ROI$ remains essentially unchanged. PRDC performs marginally below MCQ, as the nearest-neighbor-based regularization cannot effectively guide policy iterations in the highly non-stationary markets presented in the log.
%
OPAL achieves $0.66\%$ improvement in $Impr$ but its $Click$ is still declining. The reason is that its budget allocation strategy is learned through supervised learning and thus lacks proper adjustments for those undesirable budget allocation experiences. Due to limitations in policy representation capability, CEM cannot accommodate advertisers' requirements, resulting in the decrease of $CSR$ to $-5.32\%$.
Although PID adjusts bidding prices based on the current CPC dynamically, it does not take into account the inconsistent average CPC across channels and forces all channels to achieve the same level. Therefore, PID obtains a $10.15\%$ improvement in $CSR$ but poor performance in all other metrics.

\begin{figure*}[tb]
\centering
\includegraphics[width=0.85\textwidth]{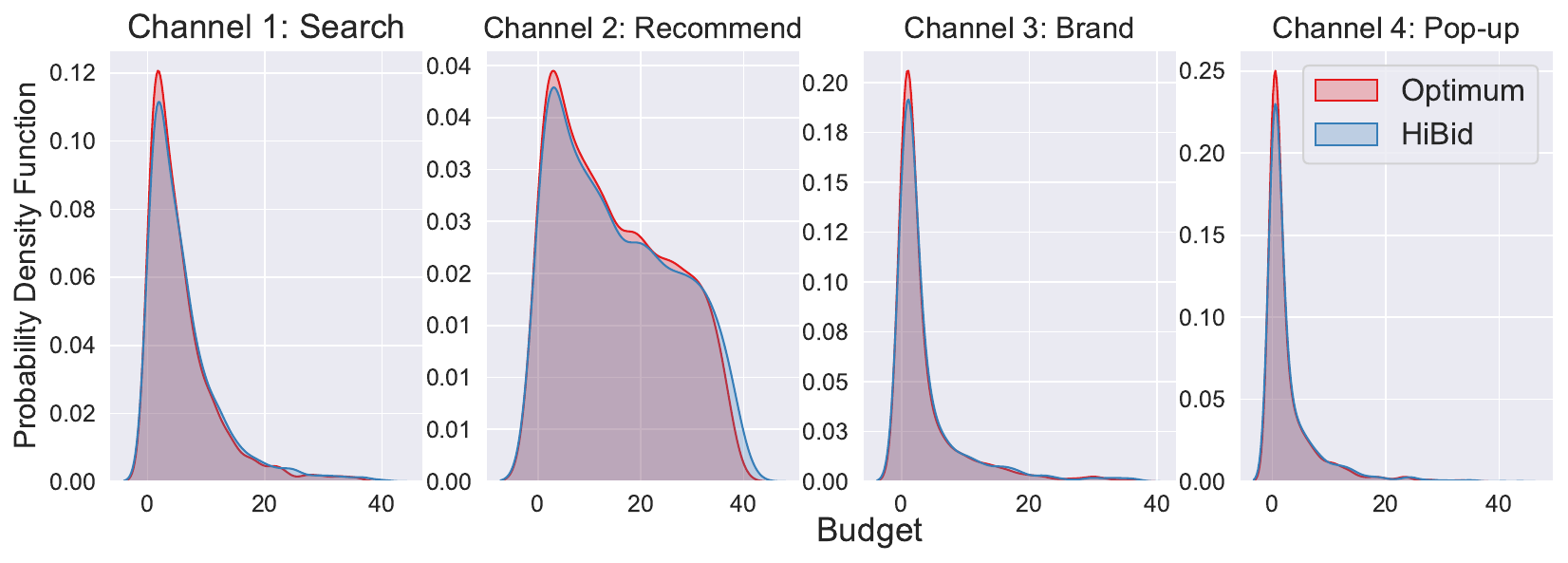}
\caption{Distribution of allocated budget between HiBid and optimum in four channels.}
\label{fig-small-scale}
\end{figure*}

\begin{table}
\centering
\setlength\tabcolsep{3pt}
\caption{Online A/B Testing}
\begin{tabular}{c|ccccc|c}
\hline
Method & $Impr$ & $Click$ & $ROI$ & $CPC$ & $CSR$ & TP999 (ms) \\ \hline
HiBid   & 0.15\%     & \textbf{11.20\%}    & \textbf{5.15}\%  & \textbf{-6.63\%}   & 9.02\%  &33.8 \\
CBRL \cite{cbrl}    & 0.21\%      & 3.15\%    & 2.16\%   & -1.89\%   & 3.79\%  &29.1 \\
MCQ \cite{mcq}     & \textbf{0.92}\%      & 2.67\%    & -0.30\%   & 0.52\%    & 1.89\%  &
28.4\\
PRDC \cite{PRDC}     & 0.75\%      & 2.39\%    & -0.56\%   & 0.50\%    & 1.91\%  & \textbf{27.9} \\
OPAL \cite{opal}   & 0.66\%     & -0.58\%    & -5.98\%   & -4.70\%   & -2.01\% &31.1 \\
CEM \cite{cem}    & -0.03\%      & 2.49\%    & -6.61\%   & -1.76\%   & -4.39\%  & - \\
PID \cite{ang2005pid}     & -0.82\%      & -14.45\%   & -2.07\%   & 0.63\%  & \textbf{9.38\%}  & - \\ \hline
\end{tabular}
\label{table_online}
\end{table}

\subsection{Online A/B Testing on Meituan Advertising Platform}
HiBid and all other six baselines are validated on Meituan advertising platform, whose system architecture is shown in Figure \ref{fig_deployment}. Multiple ad requests are generated when users use Meituan apps, and each request goes through four modules in the advertising platform: 

\textbf{Retrieval module.} Each ad request represents a browsing from a user, thus the advertising platform retrievals a set of advertisers for this ad based on the current user behavior and historical statistics. The selected advertisers have the chance to participate in the ad request auction.

\textbf{Bid module.} The advertising platform sends remote procedure call (RPC) requests to RTB systems to obtain the bidding price of each selected advertiser for this ad. Then, the RTB system constructs request-level features and advertisers-level features for online prediction and returns the bidding price to the advertising platform.

\textbf{Rank module.} When all the advertisers' bidding prices are obtained, the advertising platform ranks the bidding advertisers in descending order based on their bidding price. The advertiser with the highest bidding wins this ad and has the chance to display the ad.

\textbf{Price module.} The simulator deployed a GSP auction with CPC pricing schemes identical to the online advertising platform, which means that only if the user clicks the ad, the displayed advertiser will be charged the price of the second-highest bid in this ad auction.

Subsequently, the advertising platform will display ads to users based on the auction results, and record user feedback (e.g., whether they click on the ad or make an order) in Meituan's \textbf{Data Center}. The daily bidding logs are automatically processed and stored in the \textbf{Feature Center} for high-level and low-level training. After the offline training, the trained model is synchronized to the RTB system for bidding services. This allows the high-level planner and low-level executor of our proposed HiBid to rapidly iterate and adapt to the changing environment, leading to performance improvement in large-scale advertising systems in practice.

Online experiments are conducted using A/B testing for two weeks, and results are shown in Table \ref{table_online}. HiBid consistently outperforms all other baselines on most metrics by $11.20\%$ in $Click$ and $5.15\%$ in $ROI$ at least. Meanwhile, PID achieved a slightly higher $CSR$, but it performs poorly in other metrics. In addition, We show the TP999 (completion time for $99.9\%$ requests) for each model in Table \ref{table_online}. Due to the introduced optimal $\lambda-$selection and CPC-AS mechanism, the inference time of HiBid is only slightly increased, but it is still quite acceptable.

\subsection{Synthetic Dataset Validation}
Currently, RTB-related research efforts have not offered a publicly available dataset for performance comparison. Thus, to ease fair algorithm comparisons and code reproduction for the research community, we develop a cross-channel constrained bidding simulator to generate the synthetic dataset and make it available online. It emulates the ad display process of the advertising platform depicted in Fig. \ref{fig_deployment}. We implement the core modules as a simplified version of the advertising system, including retrieval, bid, rank, and price. Furthermore, we simulate ad requests and user feedback based on the distribution statistics obtained from Meituan's online production system, as:

\begin{table}
\centering
\caption{Synthetic Dataset Validation}
\begin{tabular}{c|ccccc}
\hline
Method & $Impr$ & $Click$ & $ROI$ & $CPC$ & $CSR$ \\ \hline
HiBid           & 0\%                    & 15.25\%                 & 6.38\%                & -8.13\%               & 14.53\%               \\
CBRL            & 0\%                    & 5.65\%                  & 1.80\%                & -6.16\%               & 8.9\%                 \\
MCQ             & 0\%                    & 3.95\%                  & 0.86\%                & -5.67\%               & 7.84\%                \\
PRDC         &    0\%                    & 3.65\%                  & 0.78\%                & -5.35\%               & 7.41\%                \\

OPAL          & 0\%                    & 1.13\%                  & -3.75\%               & -0.74\%               & 3.61\%   
\\ \hline
\end{tabular}
\label{table_synthetic}
\end{table}

\textbf{Request and advertisers simulation.} It first randomly generates a few advertisers, who are categorized into several types representing varying business conditions of advertisers from the online statistics. Each advertiser possesses a total budget, expected CPC, historical CTR, CVR, GMV, etc, which are sampled from a Gaussian distribution based on their own category. Then, the simulator initializes the total ad requests, each of which belongs to one of $P$ channels. The number of ad requests and arrival distribution of each channel are kept relatively consistent with the online platform. The simulator replays all the ad requests hand uses the base bidding strategies (i.e., CTR-based strategy that bids with predicted CTR multiply $CPC_m^{set}$) to conduct auctions.

\textbf{User feedback simulation.} After the ad auction is finished, the simulator samples the user feedback towards the displayed ads from multiple Gaussian distributions, including whether the user clicks, makes an order, and the making order amount. Then we update the advertiser's daily statistics (i.e., real-time CTR, CVR, budget consumption, the number of clicks, etc.), to simulate the real-time feature on the platform. Both user feedback and the auction process are recorded in the dataset for model offline training.

Our synthetic dataset is made publicly available online\footnote{https://drive.google.com/drive/folders/11TmSXZFtwiXhy1kyQdvEvzc5Mu-cHI7S}, including 5,000 advertisers bidding for 3,500,000 ad requests in 7 days. During the model evaluation, we first restore and replay all ad requests. Then we use the RL-based bidding strategy instead of the base strategy to participate auction and summarize the bidding results to evaluate the model performance. Results are shown in Table \ref{table_synthetic}. Since there is no external competition in the simulator, the total impressions remain unchanged. We see that Hibid consistently outperforms other baselines, improving $CSR$ while bringing more clicks for advertisers.

\subsection{HiBid Performance Verification Upon Optimality}
\label{proof}
\subsubsection{Bias Incurred during CPC-AS Derivation} 
\begin{figure}[tb]
\centering
\includegraphics[width=\columnwidth]{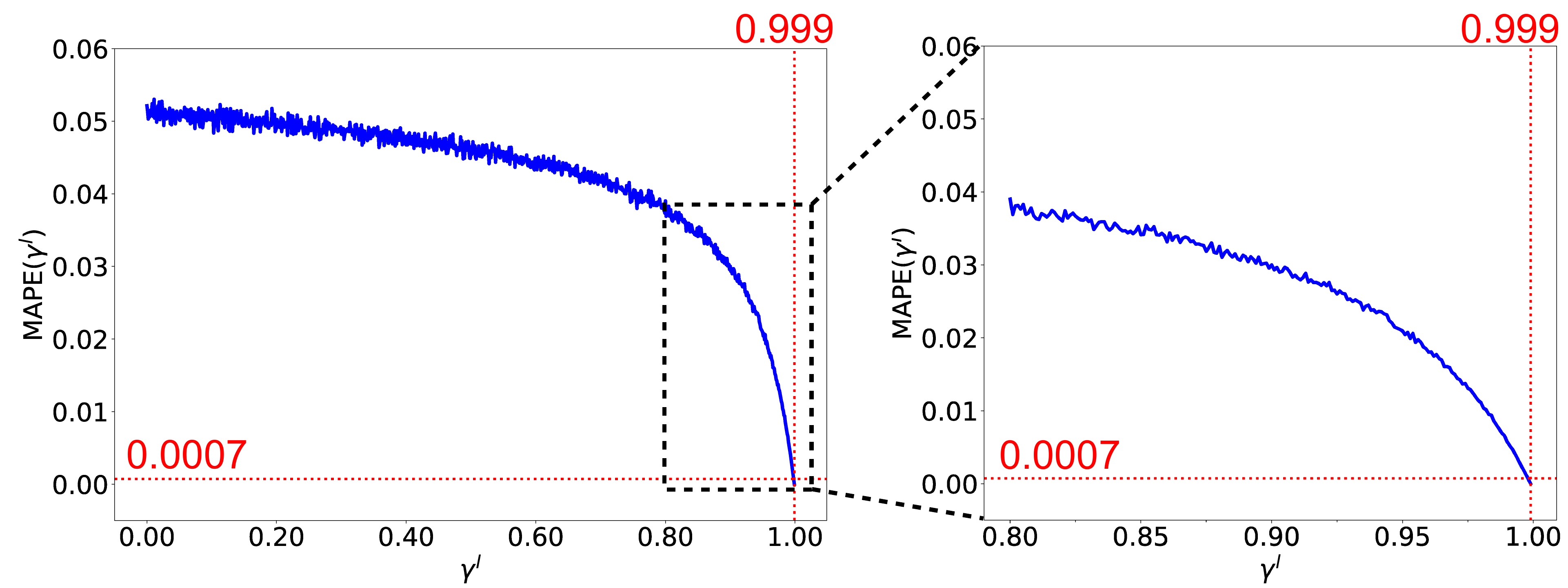}
\vspace{-3mm}
\caption{MAPE between $CPC_m^{pred}(a_{i}^l)$ and $CPC_m^{real}$ in Eqn. (\ref{mape}). }
\vspace{-1mm}
\label{fig_mape}
\end{figure}

To verify the bias between our proposed $CPC_m^{pred}(a_{i}^l)$ in Eqn. (\ref{CPC-calculate}) and $CPC_m^{real}$, we sample $13,633,825$ trajectories from the low-level dataset $\mathcal{D}^l$ and calculate the bias with different $\gamma^l\in[0,1]$. The mean absolute percentage error $\mathrm{MAPE}(\gamma^l)$ is then calculated by:
\begin{equation}
\small
\label{mape}
    \mathrm{MAPE}(\gamma^l) = \mathrm{average}\bigg(\frac{|CPC_m^{pred}(\gamma^l,a_{i}^l)-CPC_m^{real}|}{CPC_m^{real}}\bigg|_{\tau^l} \bigg),
\end{equation}
and the obtained curve is shown in Figure \ref{fig_mape}. We observe that $\mathrm{MAPE}(\gamma)$ decreases rapidly as $\gamma^l$ increases, and the difference is less than $0.001$ and thus can be ignored when $\gamma^l=0.999$. 

\subsubsection{Optimality Performance of HiBid in Small Scale Dataset}
Millions of advertisers bid for billions of ad requests every day, the solution space explodes exponentially and it is impossible to obtain the whole day's information about the incoming request, making it difficult to get an optimal solution in the considered c$^3$-bidding problem. However, the high-level budget allocation problem can be solved by optimization methods if the number of advertisers is not too large and the information of all participating advertisers is as prior.
In order to verify the performance gap between the budget allocation strategies obtained by HiBid and the optimum, we constructed a small-scale dataset containing $2,000$ advertisers and get the optimal budget allocation through a solver \cite{solver}. 
The distributions of the allocated budget are shown in Figure \ref{fig-small-scale}, and the MAPE and RMSE between HiBid and optimal are $4.76\%$ and $1.09$, respectively.
Therefore, HiBid performs well which enables it to cope with real-world complex and dynamic bidding environments.


\section{Concluding Remark}
\label{sec-con}
In this paper, we propose a hierarchical offline DRL framework HiBid for cross-channel bidding with budget allocation. HiBid introduces three contributions based on the state-of-the-art offline DRL approach MCQ: (a) auxiliary batch loss to alleviate the advertiser competition in high-quality channels, (b) $\lambda$-generalization for adaptive constrained bidding strategy in response to changing budget, and (c) CPC-guided action selection scheme for improving cross-channel CPC satisfactory ratio. Both offline experiments and online A/B testing on Meituan advertising platform show HiBid outperforms six baselines. The HiBid has been deployed online to already service tens of thousands of advertisers every day. Furthermore, some works~\cite{bidding4cloud,bidding4cloud2} have introduced an auction mechanism for resource allocation (e.g., VMs) in cloud computing. In such auction-based allocations, a hierarchical architecture can also be employed to enhance resource distribution efficiency, benefiting both the platform and its users. While the primary application of our work is ad display, the proposed hierarchical architecture offers its potential to work with other applications that require resource allocation through auctions.

\bibliographystyle{IEEEtran}
\bibliography{main}

\end{document}